\documentclass[preprint,12pt]{elsarticle}

\usepackage{times}
\usepackage{amsmath}
\usepackage{amsthm}
\usepackage{array}
\usepackage{graphicx}
\usepackage{bm}	
\usepackage[noend]{algpseudocode}
\usepackage{algorithmicx,algorithm}
\theoremstyle{definition}
\newtheorem{Thm}{\textbf{Theorem}} 
\newtheorem{Pro}[Thm]{\textbf{Proposition}} 
\newtheorem{Lem}[Thm]{\textbf{Lemma}} 

\newtheorem{Def}[Thm]{\textbf{Definition}}

\usepackage{multirow}
\usepackage{mathrsfs}
\usepackage{amssymb}
\usepackage{subfig}
\usepackage{amsthm}

\begin{document}

\begin{frontmatter}



\title{A New Learning Paradigm for Random Vector Functional-Link Network: RVFL+}



\author{Peng-Bo Zhang\corref{cor1}} 
\cortext[cor1]{Corresponding author}
\ead{pengbolong123@126.com} 

\author{Zhi-Xin Yang\corref{}} 

\ead{zxyang@um.edu.mo} 

\address{State Key Laboratory of Internet of Things for Smart City and Department of Electromechanical Engineering, Faculty of Science and Technology, University of Macau, Macau SAR, 999078}

\begin{abstract}
In school, a teacher plays an important role in various classroom teaching patterns. Likewise to this human learning activity, the learning using privileged information (LUPI) paradigm provides additional information generated by the teacher to 'teach' learning models during the training stage. Therefore, this novel learning paradigm is a typical $\emph{Teacher-Student Interaction}$ mechanism. This paper is the first to present a random vector functional link network based on the LUPI paradigm, called RVFL+. Rather than simply combining two existing approaches, the newly-derived RVFL+ fills the gap between classical randomized neural networks and the newfashioned LUPI paradigm, which offers an alternative way to train RVFL networks. Moreover, the proposed RVFL+ can perform in conjunction with the kernel trick for highly complicated nonlinear feature learning, which is termed KRVFL+. Furthermore, the statistical property of the proposed RVFL+ is investigated, and we present a sharp and high-quality generalization error bound based on the Rademacher complexity. Competitive experimental results on 14 real-world datasets illustrate the great effectiveness and efficiency of the novel RVFL+ and KRVFL+, which can achieve better generalization performance than state-of-the-art methods.

\end{abstract}


\end{frontmatter}


\section{Introduction}
\label{sec:1}

Recently, Vapnik and Vashist \cite{vapnik2009new} provided a new learning paradigm termed learning using privileged information (LUPI), which is aimed at enhancing the generalization performance of learning algorithms. Generally speaking, in classical supervised learning paradigm, the training data and test data must come from the same distribution. Although in this new learning paradigm the training data is also considered an unbiased representation for the test data, the LUPI provides a set of additional information for the training data during the training stage. The set of additional information is termed privileged information. Different from the tranditional supervised learning approaches, the LUPI based methods make use of a new kind of training data including privileged information during the training phase, but the privileged information is not available in the test stage. We note that the new learning paradigm is analogous to human learning process. In class, a teacher can provide some important and helpful information about this course for students, and these information provided by a teacher can help students acquire knowledge better. Therefore, a teacher plays an essential role in human leaning process. Likewise to the classroom teaching model, in general, the LUPI paradigm based methods can also achieve better generalization performance than traditional learning models.    

Vapnik and Vashist \cite{vapnik2009new} was the first to present a SVM algorithm with privileged information termed SVM+, which leverages the strength of the LUPI paradigm. A thorough theoretical analysis of the SVM+ was further illustrated in \cite{pechyony2010theory,vapnik2015learning}. Previous works of the LUPI paradigm focus on two aspects: solving the LUPI based algorithms efficiently and incorporating the LUPI paradigm into various learning models. This paper focuses on the latter. The newly-derived RFVL+, however, has much milder optimization constraints than the SVM+. As a result, we can obtain a closed-form solution to the new RFVL+, which naturally tackles the former.   

From the optimization perspective, the formulation of the SVM+ is a typical quadratic programming (QP) problem, and in general the QP problem can be solved by some optimization toolboxes (for example the CVX toolbox \cite{grant2008cvx}). However, it is unnatural and inconvenience to train a learning model by some optimization toolboxes in real-world applications. For this reason, it is necessary to present an efficient approach to solve it \cite{pechyony2010smo,li2016fast,pechyony2011fast}. Pechyony et. al. \cite{pechyony2010smo} presented an SMO-style optimization approach for the SVM+. Li et. al. \cite{li2016fast} further proposed two fast algorithms for linear and kernel SVM+, respectively. In addition to solving the SVM+ efficiently, the LUPI paradigm is incorporated into various learning algorithms \cite{feyereisl2014object,fouad2013incorporating,xu2015distance,lapin2014learning,sharmanska2013learning}. Feyereisl et. al. \cite{feyereisl2014object} presented a novel structured SVM for object localization, which uses attributes and segmentation masks of an object as privileged information. Fouad et. al. \cite{fouad2013incorporating} provided a generalized matrix LVQ (GMLVQ) approach based on the LUPI paradigm. In order to tackle the face verification and person re-identification problems better, Xu et. al. \cite{xu2015distance} used the depth information of RGB-D images as privileged information to present a novel distance metric learning algorithm. These existing works have confirmed the advantage of the LUPI-based learning models.

Nowadays, neural network is one of the most popular learning algorithms due to the wave of deep learning, and most of current deep learning methods are neural networks, including denoising auto-encoders (DAE) \cite{bengio2013generalized}, convolutional neural networks (CNNs) \cite{krizhevsky2012imagenet}, deep belief networks (DBNs) \cite{hinton2006fast} and long short-term memory (LSTM) \cite{hochreiter1997long}, etc. These neural network methods have achieved greatly successes in various real-world applications, including image classification and segmentation, speech recognition, natural language processing, etc. Therefore, it is very interesting to combine neural networks and the LUPI paradigm. The combined method is able to leverage the strengths of neural networks and the LUPI paradigm. The goal of this paper is to tackle this open problem and construct a bridge to link the LUPI paradigm and randomized neural networks.      

In this paper, we propose a novel random vector functional link network with privileged information called RVFL+. The random vector functional link network (RVFL) \cite{pao1995functional,pao1992functional,pao1994learning,igelnik1995stochastic} is a classical single layer feedforward neural network (SLFN), which overcomes some limitations of SLFNs including slow convergence, over-fitting and trapping in a local minimum. Although the RVFL has achieved good generalization performance in some real-world tasks \cite{zhang2016comprehensive}, in order to improve further its effectiveness, we incorporate the LUPI paradigm into the RVFL. Different from existing variants of RVFL, the RFVL+ may open a door towards alternative to the traditional learning paradigm for the RVFL in real-world tasks. In other words, the RVFL+ makes use of not only the labeled training data but also a set of additional privileged information during the training stage, which interprets the essential difference between the two learning paradigms. 

Moreover, following the kernel ridge regression \cite{vovk2013kernel,saunders1998ridge}, we further propose a kernel-based RVFL+ called KRVFL+ in order to handle highly complicated nonlinear relationships. The KRVFL+ has two major advantages over the RVFL+. On one hand, the random affine transform leading to unpredictability is eliminated in the KRVFL+. Instead, both the original and privileged features are mapped into a reproducing kernel Hilbert space (RKHS). On the other hand, the KRVFL+ no longer considers the number of enhancement nodes, which is a key factor to affect its generalization ability.  As a result, the performance of the KRVFL+ in terms of effectiveness and stability is significantly improved in most real-world tasks.       

Furthermore, we investigate the statistical property of the newly-derived RVFL+. We provide a tight generalization error bound based on the Rademacher complexity \cite{bartlett2002rademacher} for the RVFL+. Our generalization error bound benefits from the advantageous property of the Rademacher complexity. The Rademacher complexity is a commonly-used powerful tool to measure the richness of a class of real-valued functions in terms of its inputs, and thus better capture the property of distribution that generates the date. In the RVFL+, the weights and biases between the input layer and enhancement nodes are generated randomly and are fixed, the output weights are then calculated by the Moore-Penrose pseudo-inverse \cite{igelnik1995stochastic,pao1995functional} or the ridge regression \cite{bishop2006pattern}. Therefore, the RVFL+ is considered as a 'special' linear learning model. The Rademacher complexity is an ideal choice for the analysis of this type of methods, and can provide a high-quality generalization error bound in terms of its inputs. In contrasts to the previous work \cite{xu2017kernel}, we provide a more tight and general test error bound, and the novel bound is also appropriate for various versions of the RVFL including the newly-derived KRVFL+.

Last but not least, we construct some competitive experiments on 14 real-world datasets to verify the effectiveness and efficiency of the newly-derived RVFL+ and KRVFL+. The experimental results illustrate that the novel RVFL+ and KRVFL+ outperform state-of-the-art comparisons. More importantly, recent existing works have illustrated that the cascaded multi-column RVFL+ (cmcRVFL+) \cite{shi2018cascaded} and the cascaded kernel RVFL+ (cKRVFL+) \cite{dai2018transcranial} can obtain the best performance in terms of effectiveness for the single-modal neuroimaging-based diagnosis of Parkinson's disease and the transcranial sonography (TCS) based computer-aided diagnosis (CAD) of Parkinson's disease, respectively. Notice that both the RVFL+ and KRVFL+ are basic learners and play key roles in these two ensemble learning methods.     

The \emph{contributions} of this paper are summarized as follows. 
\begin{itemize}
	\item We propose a novel random vector functional link network with privileged information, called RVFL+. The RVFL+ bridges the gap between randomized neural networks and the LUPI paradigm. Different from existing variants of the RVFL, the newly-derived RVFL+ provides an alternative paradigm to train the RVFL, which is a typical multi-source feature fusion learning mechanism.   
	\item We extend the RVFL+ to the kernel version called KRVFL+, and the KRVFL+ enables handling effectively highly nonlinear relationships between high-dimensional inputs.
	\item The previous works of the LUPI focus on two aspects: deriving an efficient solver and combining the LUPI paradigm with different learning models. This paper focuses on the latter. However, from the optimization perspective, we find that the novel RVFL+ has sampler constraints than the SVM+. As a result, we can obtain a closed-form solution to the RVFL+, which naturally tackles the former.       
	\item This paper not only gives a comprehensive theoretical guarantee using the Rademacher complexity for the new RVFL+, but it also empirically verifies that the newly-derived RVFL+ and KRVFL+ outperform state-of-the-art methods on 14 real-world datasets.               
\end{itemize}  

The remainder of this paper is organized as follows. We brief the related work of the RVFL in Section \ref{sec:2}. In Section \ref{sec:3}, we briefly explain the reason that the RVFL works well for most real-world tasks, and then introduce the newly-derived RVFL+ and KRVFL+. We study the statistical property of the RVFL+ and provide a novel tight generalization error bound based on the Rademacher complexity in Section \ref{sec:4}. In Section \ref{sec:5}, we conduct several experiments on 14 real-world datasets to evaluate the proposed RVFL+ and KRVFL+. This paper concludes in Section \ref{sec:6}.

\section{Related work of Random Vector Functional-Link Networks}
\label{sec:2}
Over the last three decades, randomization based methods, including random projection \cite{liu2012texture}, random forests \cite{breiman2001random}, bagging \cite{breiman1996bagging}, stochastic configuration networks (SCN) \cite{wang2017stochastic,zhu2019further}, and random vector functional link networks (RVFL) \cite{pao1994learning}, etc., play important roles in machine learning community. We refer to \cite{cao2018review,zhang2016survey} for great surveys of the randomized neural networks. 

The RVFL presented first by Pao et. al. \cite{pao1994learning} is one of most popular single layer feedforward neural networks due to its universal approximation ability and great generalization performance. Many researchers have investigated numerous variants of the RVFL in various domains during these three decades. Chen and Wan \cite{chen1999rapid} presented two novel algorithms for the functional-link network in order to calculate efficiently the optimal weights and update the weights on-the-fly, receptively. Chen \cite{chen1996rapid} presented a novel single-hidden layer neural
network structure, which can rapidly calculate the optimal weights. A new RVFL was presented by Patra et. al. \cite{patra1999identification} for nonlinear dynamic systems. In addition to these early studies, the RVFL has gained a huge attention from more and more researchers in recent years. Cui et. al. \cite{cui2018received} presented a novel system based on a RVFL to address the indoor positioning problem. Zhang et. al. \cite{zhang2019unsupervised} presented a new sparse pre-trained RVFL method (SP-RVFL for short) in order to address classification problems. Scardapane et. al. \cite{scardapane2018bayesian} proposed a new RVFL with full Bayesian inference for robust data modeling. The parsimonious random vector functional link network (pRVFLN) was presented in \cite{pratama2018parsimonious} for data stream problems, which overcomes the limitation of the original RVFL in such cases. Xu et. al. \cite{xu2017kernel} built an effective spatiotemporal model based on a kernel-based RVFL in order to forecast the distribution of the temperature. Following the basic idea of the RVFL, a broad learning system proposed by Chen and Liu \cite{chen2018broad} provides an alternative mode to design the architecture of learning algorithms in this big data era \cite{chen2014data}. Therefore, we notice that the RVFL is an exceedingly  powerful model that is worth understanding, promoting, and developing.   

\section{Random Vector Functional-Link Networks with Privileged Information}
\label{sec:3}
\subsection{Preliminaries}
\begin{figure}[htpb]
	\centering
	\includegraphics[width=\textwidth]{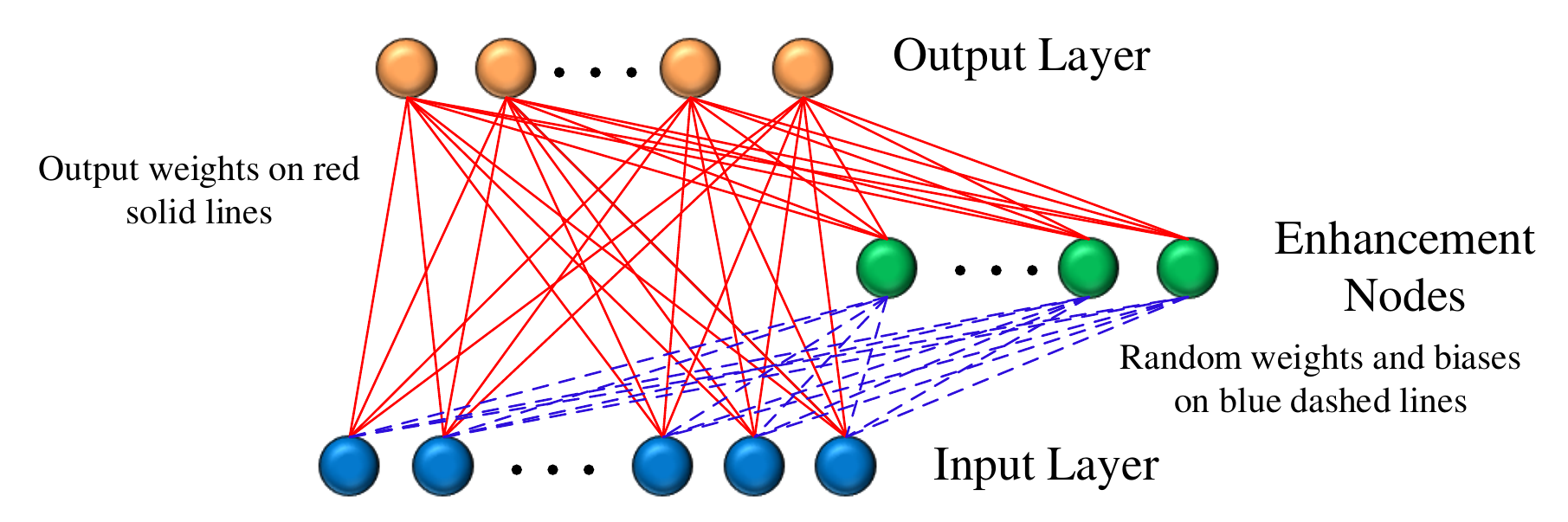}
	\caption{The architecture of the RVFL network}
	\label{fig:1}
\end{figure}

The RVFL network is a classical single layer feedforward neural network, and the architecture of the RVFL is shown in Figure \ref{fig:1}. The RVFL initializes randomly all weights and biases between the input layer and enhancement nodes, and then these parameters are fixed and do not need to be tuned during the training stage. The output weights on red solid lines in Figure \ref{fig:1} can be calculated by the Moore-Penrose pseudo-inverse \cite{igelnik1995stochastic,pao1995functional} or the ridge regression \cite{bishop2006pattern}. Moreover, the direct link between the input layer and the output layer is an effective and simple regularization technique preventing RVFL networks from overfitting.

Given a set of labeled data $\{(\bm{x}_i, \bm{y}_i)|\bm{x}_i \in \mathbb{R}^n, \bm{y}_i \in \mathbb{R}^m, i = 1, \dots, N\}$, a RVFL network with $P$ enhancement nodes can be formulated as 
\begin{align}
\label{eq:1}
\bm{H}\bm{w} = \bm{Y}
\end{align}
where $\bm{w}$ is an output weight vector, $\bm{H}$ is a concatenated matrix combining input data and outputs from the enhancement nodes and $\bm{Y}$ is a label matrix. $\bm{H}$ and $\bm{w}$ are shown as  
\begin{align}
\label{eq:2}
\bm{H} = \left[\bm{H}_1 \quad \bm{H}_2\right] \nonumber \\ \bm{H}_1 = \left[\begin{array}{ccc}  
\bm{x}_{11} & \cdots & \bm{x}_{1n} \\
\vdots & \ddots & \vdots \\ 
\bm{x}_{N1} & \cdots & \bm{x}_{Nn}\\
\end{array}\right]
\nonumber \\
\bm{H}_2 = \left[\begin{array}{ccc}  
G(\bm{a}_1\cdot\bm{x}_1+b_1) & \cdots & G(\bm{a}_P\cdot\bm{x}_1+b_P) \\
\vdots & \ddots & \vdots \\ 
G(\bm{a}_1\cdot\bm{a}_N+b_1) & \cdots & G(\bm{a}_P\cdot\bm{a}_N+b_P)\\
\end{array}\right]
\end{align}
\begin{align}
\bm{w} = \left[\begin{array}{c}  
\bm{w}_1^T \\
\vdots \\ 
\bm{w}_{(n+P)}^T\\
\end{array}\right]
\end{align} 

In (\ref{eq:2}), $\bm{a}_j$ and $b_j$ ($j = 1, \dots, P$) are the weight and bias between the input layer and enhancement nodes. According to \cite{zhang2016comprehensive}, all weights and biases are from a uniform distribution within $[-u, u]$ and $[0, u]$, respectively, where $u$ is a positive user-defined parameter. This reasonable initialization can overcome the limitation of the RVFL. In other words, the RVFL with random weights and biases uniformly chosen from $[-1, 1]$ and $[0, 1]$ respectively may fail or not generate an optimal solution. $G(\cdot)$ is a nonlinear activation function such as sigmoid, tanh, rbf, etc.

From (\ref{eq:1}), we can calculate directly the output weights $\bm{w}$ by the Moore-Penrose pseudo-inverse \cite{igelnik1995stochastic,pao1995functional} or the ridge regression \cite{bishop2006pattern}, which are shown in (\ref{eq:3}) and (\ref{eq:4}) respectively.
\begin{align}
\label{eq:3}
\bm{w} = \bm{H}^{\dagger}\bm{Y}
\end{align}
\begin{align}
\label{eq:4}
\bm{w} = (\bm{H}^T\bm{H}+\frac{\bm{I}}{C})^{-1}\bm{H}^T\bm{Y}
\end{align}
where $\dagger$ is the Moore-Penrose pseudo-inverse, $\bm{I}$ is an identity matrix and $C$ is a trading-off parameter. 

The model structure of the RVFL is so simple, why does the RVFL work well for most tasks? Giryes et. al. \cite{giryes2016deep} give a possible theoretical explanation for this open problem. To provide this explanation, first of all, Giryes et. al. reveal the essence of training in learning models. Generally speaking, the angles of instances between different classes are larger than ones in the same class \cite{wolf2003learning}. Therefore, from a geometric point of view, the role of training stage is to penalize the angles between instances from different classes more than ones in the same class \cite{giryes2016deep}. Moreover, in this big data era, due to the highly complicated model architecture and an increasing number of trainable parameters, it is quick difficult to tune all parameters in learning process, which needs extremely high computational cost. To tackle this problem, randomization is an ideal choice for some learning models, resulting in cheaper computational cost. A great random initialization allows the learning model to be a universal one before training. Therefore, many researchers have investigated various initializations \cite{glorot2010understanding, he2015delving} in order to train networks better.   Now we revisit the RVFL. The RVFL uses the hybrid strategy to train the entire network. In the RVFL, the random initial parameters between the input layer and the enhancement nodes solve well inputs having distinguishable angles, and the turned output weights further deal with the remaining instances.

\subsection{RVFL+ for Baniry, Multiclass Classification and Regression}
In feedforward neural networks having small empirical error, Bartlett \cite{bartlett1998sample} illustrates that the smaller the norm of weights are, the greater generalization performance is. From the optimization perspective, we note that the basic idea of the RVFL network is to minimize the training error $\|\bm{Y}-\bm{H}\bm{w}\|_2^2$ and the output weights $\|\bm{w}\|^2_2$ simultaneously, which emphasizes the combination of the least square loss function and the Tikhonov regularization. As a consequence, the RVFL can achieve a great generalization performance. In order to incorporate the LUPI paradigm, following the dual version of the
ridge regression approach \cite{saunders1998ridge}, we formulate the RVFL as  
\begin{align}
\label{eq:5}
&\min_{\bm{w},\bm{\zeta}} \frac{1}{2} \|\bm{w}\|_2^2+\frac{C}{2} \sum_{i=1}^{N} \bm{\zeta}_i^2 \nonumber \\
&s.t. \quad \bm{h}(\bm{x}_i)\bm{w} = \bm{y}_i-\bm{\zeta}_i, \forall 1 \leq i \leq N.
\end{align}
where $\bm{h}(\bm{x}_i)$ is a 'combined' features vector, including the original input features vector $\bm{h}_1$ and the output features vector $\bm{h}_2$ from the enhancement nodes. We define $\bm{h}(\bm{x}_i)$ as the enhanced layer output vector. $\bm{\zeta}_i = [\zeta_{i1}, \dots, \zeta_{im}]^T$ is a training error vector with $m$ output nodes, and $\bm{y}_i$ is a one-hot label vector. 

Following the relationship between SVM \cite{cortes1995support} and LS-SVM \cite{suykens1999least}, we minimize $\bm{\zeta}_i$ in (\ref{eq:5}) instead of $\bm{\zeta}_i^2$ in order to incorporate the LUPI paradigm easily. Therefore, we have
\begin{align}
&\min_{\bm{w},\bm{\zeta}} \frac{1}{2} \|\bm{w}\|_2^2+C \sum_{i=1}^{N} \bm{\zeta}_i \nonumber \\
&s.t. \quad \bm{h}(\bm{x}_i)\bm{w} = \bm{y}_i-\bm{\zeta}_i, \forall 1 \leq i \leq N.
\end{align}

Now we use the new learning  paradigm to train the RVFL network. Given a set of additional privileged information $\{\tilde{\bm{x}}_i \in \mathbb{R}^d, i = 1, \dots, N\}$ in the training process, the training data becomes $\{(\bm{x}_i, \tilde{\bm{x}}_i, \bm{y}_i)|\bm{x}_i \in \mathbb{R}^n, \tilde{\bm{x}}_i \in \mathbb{R}^d, \bm{y}_i \in \mathbb{R}^m, i = 1, \dots, N\}$. In this new training set, $\bm{x}_i \in X$ is the original feature and the privileged feature $\tilde{\bm{x}}_i \in \tilde{X}$ belongs to the privileged feature space $\tilde{X}$. In general, the privileged feature space $\tilde{X}$ is different from the original feature space $X$.   

Following the formula of the SVM+ in \cite{vapnik2009new}, we can write the RVFL+ as  
\begin{align}
\label{eq8}
&\min_{\bm{w},\tilde{\bm{w}},\bm{\zeta}} \frac{1}{2} \|\bm{w}\|_2^2+\frac{\gamma}{2} \|\tilde{\bm{w}}\|_2^2+C \sum_{i=1}^{N} \zeta_i(\tilde{\bm{w}},\tilde{\bm{h}}(\tilde{\bm{x}}_i)) \nonumber \\
&s.t. \quad \bm{h}(\bm{x}_i)\bm{w} = \bm{y}_i-\zeta_i(\tilde{\bm{w}},\tilde{\bm{h}}(\tilde{\bm{x}}_i)), \forall 1 \leq i \leq N.
\end{align}
where $\gamma$ is a regularization coefficient. Likewise to $\bm{h}(\bm{x}_i)$, $\tilde{\bm{h}}(\tilde{\bm{x}}_i)$ is also an enhanced layer output vector corresponding to the privileged feature $\tilde{\bm{x}}_i$, which can be calculated in the same fashion. $\zeta_i(\tilde{\bm{w}},\tilde{\bm{h}}(\tilde{\bm{x}}_i))$ is the correcting function (or slack function) in the privileged feature space, and $\tilde{\bm{w}}$ is an output weight vector for the correcting function.
\begin{align}
\label{eq9}
\zeta_i(\tilde{\bm{w}},\tilde{\bm{h}}(\tilde{\bm{x}}_i)) = \tilde{\bm{h}}(\tilde{\bm{x}}_i)\tilde{\bm{w}}
\end{align}

Substituting (\ref{eq9}) into (\ref{eq8}), we have the primal form of the RVFL+ as follows.
\begin{align}
\label{eq10}
&\min_{\bm{w},\tilde{\bm{w}},\bm{\zeta}} \frac{1}{2} \|\bm{w}\|_2^2+\frac{\gamma}{2} \|\tilde{\bm{w}}\|_2^2+C \sum_{i=1}^{N} \tilde{\bm{h}}(\tilde{\bm{x}}_i)\tilde{\bm{w}} \nonumber \\
&s.t. \quad \bm{h}(\bm{x}_i)\bm{w} = \bm{y}_i-\tilde{\bm{h}}(\tilde{\bm{x}}_i)\tilde{\bm{w}}, \forall 1 \leq i \leq N.
\end{align}

From (\ref{eq10}), we note that the RVFL+ minimizes the objective function over both $\bm{w}$ and $\tilde{\bm{w}}$. Therefore, not only the original features but also the privileged information determine meanwhile the separating hyperplane of the RVFL+ during the training stage.

Moreover, in contrasts to the primal form of the SVM+ in \cite{vapnik2009new}, the correcting function of the RVFL+ is either positive or negative. In other words, the RVFL+ does not consider a group of constraints $\zeta_i(\tilde{\bm{w}},b, \phi(\tilde{\bm{x_i}})) \geq 0 $ $(i = 1, \dots, N)$. As a result, the number of constraints in the RVFL+ is at least $N$ less than that of the SVM+ for the binary classification, which results that the RVFL+ has much milder optimization constraints than the SVM+.

Furthermore, in order to address the optimization problem in (\ref{eq10}), we construct the Lagrangian function $\mathscr{L}(\bm{w},\tilde{\bm{w}},\bm{\lambda})$ as   
\begin{align}
\label{eq11}
\min_{\bm{w},\tilde{\bm{w}},\bm{\lambda}} \frac{1}{2} \|\bm{w}\|_2^2+&\frac{\gamma}{2} \|\tilde{\bm{w}}\|_2^2+C \sum_{i=1}^{N} \tilde{\bm{h}}(\tilde{\bm{x}}_i)\tilde{\bm{w}} \nonumber \\ &- \sum_{i=1}^{N} \bm{\lambda}_i(\bm{h}(\bm{x}_i)\bm{w}- \bm{y}_i+\tilde{\bm{h}}(\tilde{\bm{x}}_i)\tilde{\bm{w}})
\end{align}
where $\bm{\lambda} = [\bm{\lambda}_1,\dots,\bm{\lambda}_N]^T$ are Lagrange multipliers.

To find solutions, we use the KKT condition to calculate the saddle points of the Lagrangian function $\mathscr{L}(\bm{w},\tilde{\bm{w}},\bm{\lambda})$ with respect to $\bm{w}$, $\tilde{\bm{w}}$ and $\bm{\lambda}$.
\begin{align}
\frac{\partial\mathscr{L}(\bm{w},\tilde{\bm{w}},\bm{\lambda})}{\partial \bm{w}} &= 0 \longrightarrow \bm{w} = \bm{H}^T\bm{\lambda} \label{eq:7} \\
\frac{\partial\mathscr{L}(\bm{w},\tilde{\bm{w}},\bm{\lambda})}{\partial \tilde{\bm{w}}} &= 0 \longrightarrow \tilde{\bm{w}} = \frac{1}{\gamma} (\tilde{\bm{H}}^T\bm{\lambda}-\tilde{\bm{H}}^TC\bm{1})\label{eq:8} \\
\frac{\partial\mathscr{L}(\bm{w},\tilde{\bm{w}},\bm{\lambda})}{\partial \bm{\lambda}_i} &= 0 \longrightarrow \bm{h}(\bm{x}_i)\bm{w}- \bm{y}_i+\tilde{\bm{h}}(\tilde{\bm{x}}_i)\tilde{\bm{w}} = 0 \nonumber \\ \forall 1 \leq i \leq N. \label{eq:9}
\end{align}
where $\bm{1} \in \mathscr{R}^{N \times m}$ is an identify matrix. $\tilde{\bm{H}}$ is also a concatenated output matrix from the enhancement nodes, which corresponds to the privileged features.  

Substituting (\ref{eq:7}) and (\ref{eq:8}) into (\ref{eq:9}), we have
\begin{align}
\label{eq:10}
\bm{H}\bm{H}^T\bm{\lambda} + \frac{1}{\gamma}\tilde{\bm{H}}\tilde{\bm{H}}^T(\bm{\lambda}-C\bm{1}) = \bm{Y}
\end{align}

We can further reformulate (\ref{eq:10}) as
\begin{align}
\label{eq:11}
(\bm{H}\bm{H}^T + \frac{1}{\gamma}\tilde{\bm{H}}\tilde{\bm{H}}^T)\bm{\lambda} = \bm{Y} - \frac{C\bm{1}}{\gamma}\tilde{\bm{H}}\tilde{\bm{H}}^T
\end{align}

Combining (\ref{eq:7}) and (\ref{eq:11}), we obtain the closed-form solution to the RVFL+ as follows.  
\begin{align}
\bm{w} = \bm{H}^T(\bm{H}\bm{H}^T + \frac{1}{\gamma}\tilde{\bm{H}}\tilde{\bm{H}}^T)^{-1} (\bm{Y} - \frac{C\bm{1}}{\gamma}\tilde{\bm{H}}\tilde{\bm{H}}^T)
\end{align}

According to the ridge regression \cite{bishop2006pattern}, we also impose an additional term $\frac{\bm{I}}{C}$ in order to avoid singularity and guarantee the stability of the RVFL+. As a result, we can achieve the aftermost closed-form solution to the RVFL+ as 
\begin{align}
\label{eq:12}
\bm{w} = \bm{H}^T(\bm{H}\bm{H}^T + \frac{1}{\gamma}\tilde{\bm{H}}\tilde{\bm{H}^T}+\frac{\bm{I}}{C})^{-1} (\bm{Y} - \frac{C\bm{1}}{\gamma}\tilde{\bm{H}}\tilde{\bm{H}}^T)
\end{align}

Consequently, the output function of the RVFL+ is defined as
\begin{align}
\label{eq:13}
&f(\bm{x}) = \bm{h}(\bm{x})\bm{w} = \nonumber\\ &\bm{h}(\bm{x})\bm{H}^T(\bm{H}\bm{H}^T + \frac{1}{\gamma}\tilde{\bm{H}}\tilde{\bm{H}}^T+\frac{\bm{I}}{C})^{-1} (\bm{Y} - \frac{C\bm{1}}{\gamma}\tilde{\bm{H}}\tilde{\bm{H}}^T)
\end{align}

In addition, we can obtain straightforwardly the output function $f_{test}(\bm{z}) = \bm{h}(\bm{z})\bm{w}$ in the test stage, when using the test data $\bm{z}$ instead of the training data $\bm{x}$.

The pseudo-code of the RVFL+ is summarized in Algorithm \ref{alg:1}.
\begin{algorithm}[ht]
	\caption{Random Vector Functional Link Networks with Privileged Information: RVFL+}
	\label{alg:1}
	\hspace*{0.02in} {\bf Input:}
	A set of training data $\{(\bm{x}_i, \tilde{\bm{x}}_i, \bm{y}_i)| \bm{x}_i \in \mathbb{R}^n, \tilde{\bm{x}}_i \in \mathbb{R}^d, \bm{y}_i \in \mathbb{R}^m, \forall 1 \leq i \leq N \}$; a nonlinear activation function $G(\cdot)$; the number of enhancement nodes $P$; the user-specified coefficients $C$, $\gamma$ and $u$. \\
	\hspace*{0.02in} {\bf Output:} 
	the output weight vector $\bm{w}$  
	\begin{algorithmic}[1]
		\State Initialize randomly the weights $\bm{a}$ and biases $\bm{b}$ between the input layer and the enhancement nodes from a uniform distribution within $[-u, u]$ and $[0, u]$, respectively, and then, these generated weights and biases are fixed;
		\State Calculate the output matrix $\bm{H}$ using (\ref{eq:2}); 
		\State Calculate the output matrix $\tilde{\bm{H}}$ in the same fashion;
		\State Calculate the output weight vector $\bm{w}$ of RVFL+ using (\ref{eq:12});
		\State \Return the output weight vector $\bm{w}$
	\end{algorithmic}
\end{algorithm}

For the SVM+ \cite{vapnik2009new}, the major challenge in terms of optimization arises from a group of constraints $\sum_{i=1}^{N} (\varphi_i + \psi_i -C) = 0$, where $\varphi_i$ and $\psi_i$ are Lagrange multipliers in the SVM+. Since these two sets of Lagrange multipliers need to be considered at the same time, which is difficult to solve it \cite{pechyony2010smo}. In addition to the above constraints, the RVFL+ also eliminates the other constraints $\sum_{i=1}^{N} \varphi_i y_i =0$ in the SVM+. As a result, the RVFL+ has much simpler optimization constraints than the SVM+, and can obtain a closed-form solution.    

Moreover, the RVFL+ is an unified learning model for all binary, multiclass classification and regression, and the output function in (\ref{eq:13}) can be straightforwardly applied in all three tasks. 
\begin{itemize}
	\item Binary Classification: The predicted label of the test sample is determined by 
	\begin{align}
	\hat{y} = sign(f_{test}(\bm{z}))
	\end{align}  
	\item Multiclass Classification: We adopt the one-vs.-all (OvA) strategy to determine the predicted label in the multiclass classification. Let $f_{test}^k(\bm{z})$ be the output function of the $k$-th output nodes. The predicted label of the test sample is determined by 
	\begin{align}
	\hat{y} = \arg\max_{k \in 1, \dots, m} f_{test}^k(\bm{z})
	\end{align}
	\item Regression: The predicted value is equal to the output function $f_{test}(\bm{z})$ of the RVFL+
	\begin{align}
	\hat{y} = f_{test}(\bm{z})
	\end{align}
\end{itemize}

\subsection{Kernel Extension}

In this section, we propose a kernel based random vector functional-link network with privileged information (KRVFL+ for short). There are two major advantages over the RVFL+. The KRVFL+ no longer considers the number of enhancement nodes, instead, the KRVFL+ maps the input data into a reproducing kernel Hilbert space (RKHS) in order to construct a Mercer kernel. On the other hand, the KRVFL+ is much more robust than the RVFL+. Since the KRVFL+ does not perform the random affine transformation between the input layer and enhancement nodes, and the enhanced layer output matrix is fixed when using kernel tricks. As a consequence, the generalization performance of the KRVFL+ can be improved in terms of effectiveness and stability in most real-world tasks. 

Likewise to \cite{xu2017kernel}, we reformulate (\ref{eq:13}) as
\begin{align}
\label{eq:14}
&\bm{f}(\bm{x}) = \left[\bm{h}_1 \quad \bm{h}_2 \right] \left[\begin{array}{c} \bm{H}_1^T \\ \quad\\ \bm{H}_2^T \end{array}\right] \times \nonumber\\ &(\left[\bm{H}_1 \quad \bm{H}_2 \right] \left[\begin{array}{c} \bm{H}_1^T \\ \quad\\ \bm{H}_2^T \end{array}\right] + \frac{1}{\gamma}\left[\tilde{\bm{H}}_1 \quad \tilde{\bm{H}}_2 \right] \left[\begin{array}{c} \tilde{\bm{H}}_1^T \\ \quad\\ \tilde{\bm{H}}_2^T \end{array}\right]+\frac{\bm{I}}{C})^{-1} \times \nonumber\\ &(\bm{Y} - \frac{C\bm{1}}{\gamma}\left[\tilde{\bm{H}}_1 \quad \tilde{\bm{H}}_2 \right] \left[\begin{array}{c} \tilde{\bm{H}}_1^T \\ \quad\\ \tilde{\bm{H}}_2^T \end{array}\right])
\end{align} 

We can further simplify (\ref{eq:14}) as 
\begin{align}
\label{eq:15}
\bm{f}(\bm{x}) &= (\bm{h}_1\bm{H}_1^T + \bm{h}_2\bm{H}_2^T) \times \nonumber \\ &(\bm{H}_1\bm{H}_1^T + \bm{H}_2\bm{H}_2^T+ \tilde{\bm{H}}_1\tilde{\bm{H}}_1^T + \tilde{\bm{H}}_2\tilde{\bm{H}}_2^T + \frac{\bm{I}}{C})^{-1} \times \nonumber \\ &(\bm{Y} - \frac{C\bm{1}}{\gamma}(\tilde{\bm{H}}_1\tilde{\bm{H}}_1^T + \tilde{\bm{H}}_2\tilde{\bm{H}}_2^T)) 
\end{align}

We define the kernel matrices for the KRVFL+ as 
\begin{align}
\label{eq:16}
\bm{\Omega}_1 = \bm{H}_1\bm{H}_1^T: \Omega_{1ij} = K_1(\bm{x}_i, \bm{x}_j) \nonumber \\
\bm{\Omega}_2 = \bm{H}_2\bm{H}_2^T: \Omega_{2ij} = K_2(\bm{x}_i, \bm{x}_j) \nonumber\\
\tilde{\bm{\Omega}}_1 = \tilde{\bm{H}}_1\tilde{\bm{H}}_1^T: \tilde{\Omega}_{1ij} = \tilde{K}_1(\tilde{\bm{x}}_i, \tilde{\bm{x}}_j) \nonumber \\
\tilde{\bm{\Omega}}_2 = \tilde{\bm{H}}_2\tilde{\bm{H}}_2^T: \tilde{\Omega}_{2ij} = \tilde{K}_2(\tilde{\bm{x}}_i, \tilde{\bm{x}}_j)
\end{align}
where $K_1$ and $\tilde{K}_1$ are linear kernels, as well as $K_2$ and $\tilde{K}_2$ are general Mercer kernels such as Gaussian kernel, polynomial kernel, and wavelet kernel, etc. 

Substituting (\ref{eq:16}) into (\ref{eq:15}), we have
\begin{align}
\bm{f}_{kernel}(\bm{x}) =  \left(\left[\begin{array}{c}  
\bm{K}_1(\bm{x},\bm{x}_1) \\
\vdots \\ 
\bm{K}_1(\bm{x},\bm{x}_N)\\
\end{array}\right] + \left[\begin{array}{c}  
\bm{K}_2(\bm{x},\bm{x}_1) \\ 
\vdots \\ 
\bm{K}_2(\bm{x},\bm{x}_N)\\
\end{array}\right]\right) \times \nonumber\\  (\bm{\Omega}_1 + \bm{\Omega}_2+ \frac{1}{\gamma}(\tilde{\bm{\Omega}}_1 + \tilde{\bm{\Omega}}_2) + \frac{\bm{I}}{C})^{-1}(\bm{Y} - \frac{C\bm{1}}{\gamma}(\tilde{\bm{\Omega}}_1 + \tilde{\bm{\Omega}}_2))
\end{align}
where the output weight vector $\bm{w}_{kernel}$ of the KRVFL+ is defined as 
\begin{align}
\label{eq17}
\bm{w}_{kernel} = (\bm{\Omega}_1 + \bm{\Omega}_2+ \frac{1}{\gamma}(\tilde{\bm{\Omega}}_1 +& \tilde{\bm{\Omega}}_2) + \frac{\bm{I}}{C})^{-1} \nonumber \\ &\times (\bm{Y} - \frac{C\bm{1}}{\gamma}(\tilde{\bm{\Omega}}_1 + \tilde{\bm{\Omega}}_2))
\end{align} 
According to the property of the Mercer kernel \cite{bishop2006pattern}, we can achieve the aftermost formula of the KRVFL+ as follows.
\begin{align}
\label{eq18}
\bm{f}_{kernel}(\bm{x}) =  \left(\left[\begin{array}{c}  
\bm{K}_1(\bm{x},\bm{x}_1) \\
\vdots \\ 
\bm{K}_1(\bm{x},\bm{x}_N)\\
\end{array}\right] + \left[\begin{array}{c}  
\bm{K}_2(\bm{x},\bm{x}_1) \\ 
\vdots \\ 
\bm{K}_2(\bm{x},\bm{x}_N)\\
\end{array}\right]\right)  \times \nonumber\\  (\bm{\Omega} + \frac{1}{\gamma}\tilde{\bm{\Omega}} + \frac{\bm{I}}{C})^{-1}(\bm{Y} - \frac{C\bm{1}}{\gamma}\tilde{\bm{\Omega}})
\end{align}
where $\bm{\Omega} = \bm{\Omega}_1 + \bm{\Omega}_2$ and $\tilde{\bm{\Omega}} = \tilde{\bm{\Omega}}_1 + \tilde{\bm{\Omega}}_2$.

Likewise to the RVFL+, the KRVFL+ can straightforwardly calculate the output function $f_{kernel-test}(\bm{z})$ in the test stage, which is shown as
\begin{align}
\bm{f}_{kernel-test}(\bm{z}) =  \left(\left[\begin{array}{c}  
\bm{K}_1(\bm{z},\bm{x}_1) \\
\vdots \\ 
\bm{K}_1(\bm{z},\bm{x}_N)\\
\end{array}\right] + \left[\begin{array}{c}  
\bm{K}_2(\bm{z},\bm{x}_1) \\ 
\vdots \\ 
\bm{K}_2(\bm{z},\bm{x}_N)\\
\end{array}\right]\right)  \times \nonumber\\  (\bm{\Omega} + \frac{1}{\gamma}\tilde{\bm{\Omega}} + \frac{\bm{I}}{C})^{-1}(\bm{Y} - \frac{C\bm{1}}{\gamma}\tilde{\bm{\Omega}})
\end{align}

The pseudo-code of the KRVFL+ is summarized in Algorithm \ref{alg:2}.
\begin{algorithm}[ht]
	\caption{Kernel based Random Vector Functional Link Networks with Privileged Information: KRVFL+}
	\label{alg:2}
	\hspace*{0.02in} {\bf Input:}
	A set of training data $\{(\bm{x}_i, \tilde{\bm{x}}_i, \bm{y}_i)| \bm{x}_i \in \mathbb{R}^n, \tilde{\bm{x}}_i \in \mathbb{R}^d, \bm{y}_i \in \mathbb{R}^m, \forall 1 \leq i \leq N \}$; a Mercer kernel function (for example Gaussian kernel); the user-specified coefficients $C$ and $\gamma$. \\
	\hspace*{0.02in} {\bf Output:} 
	the output weight vector $\bm{w}_{kernel}$.  
	\begin{algorithmic}[1]
		\State Calculate the linear kernel functions $\bm{\Omega}_1$ and $\tilde{\bm{\Omega}}_1$ corresponding to $\bm{x}$ and $\tilde{\bm{x}}$, respectively;
		\State Calculate the general Mercer kernel functions (for example Gaussian kernel) $\bm{\Omega}_2$ and $\tilde{\bm{\Omega}}_2$ corresponding to $\bm{x}$ and $\tilde{\bm{x}}$, respectively; 
		\State Calculate the output weight vector $\bm{w}_{kernel}$ of KRVFL+ using (\ref{eq17});
		\State \Return the output weight vector $\bm{w}_{kernel}$.
	\end{algorithmic}
\end{algorithm}

\section{Theoretical analysis of RVFL+}
\label{sec:4}
In this section, we investigate the statistical property of the RVFL+, and only consider the binary classification for simplicity. Following the Rademacher complexity, we provide a tight generalization error bound for the RVFL+. We assume all sets considered in this paper are measurable. First of all, we give the following fact, which is a necessary condition for the Rademacher complexity.

\begin{Pro}
	\label{thm:1}
	The loss function $\ell$ in the RVFL+ satisfies Lipschitz continuity, and there exists a positive Lipschitz constant $K$, that is, for $\forall$ $\bm{x}$, $\bm{y}$ $\in$ $R^n$  
	\begin{align}
	\|\ell(\bm{x})-\ell(\bm{y})\| \leq K\|\bm{x}-\bm{y}\|
	\end{align}
	where $\|\cdot\|$ is a norm function.
	\begin{proof}
		It is straightforward that the loss function $\ell$ in the RVFL+ is a norm function, which must satisfy Lipschitz continuity.
	\end{proof}
\end{Pro}

The novel generalization error bound is dependent on the Rademacher complexity, and thus, we give the definition of the Rademacher complexity.
\begin{Def} \cite{bartlett2002rademacher}
	Given a set of i.i.d. samples $\bm{\mu}_1, \dots, \bm{\mu}_M$, where $\bm{\mu}_i \in U, \forall 1 \leq i \leq M$. Let $\mathscr{F}$ be a family of functions mapping from the space $U$ to an output space. Then, we define the Rademacher complexity of $\mathscr{F}$ as 
	\begin{align}
	\mathscr{R}_M(\mathscr{F}) = \mathbb{E}\left[\sup_{f\in \mathscr{F}} \frac{1}{M} \sum_{i=1}^{M} f(\bm{\mu}_i)\epsilon_i \right]
	\end{align}
	where $\epsilon_i, i = 1, \dots, M$ are i.i.d $\{\pm1\}$-valued Bernoulli random variables drawn at equal probability and are independent of inputs.
\end{Def}

According to \cite{bartlett2002rademacher}, we illustrate the general mathematical formula of the generalization error bound based on the Rademacher complexity in the following theorem. 
\begin{Thm} \cite{bartlett2002rademacher}
	\label{thm:2}
	Let $\mathfrak{L}(f)$ and $\hat{\mathfrak{L}}(f)$ be the generalization error bound and the empirical error bound, respectively, and let $\mathscr{F}$ be a family of functions. For a Lipschitz continuous loss function $\ell$ bounded by $c$, with probability at least $1-\delta$ ($\delta \in (0,1)$) over the samples with $M$, for all $ f \in \mathscr{F}$, we have
	\begin{align}
	\mathfrak{L}(f) \leq \hat{\mathfrak{L}}(f) + 2K\mathscr{R}_M(\mathscr{F})+c\sqrt{\frac{ln(1/\delta)}{2M}}
	\end{align}
\end{Thm}

From Theorem \ref{thm:2}, we note that the generalization error is bounded by the Rademacher complexity. Therefore, according to \cite{kakade2009complexity}, we give a Rademacher complexity of the proposed RVFL+, which serves to bound the generalization error of the RVFL+. The following Lemma is helpful in bounding the Rademacher complexity. 
\begin{Lem}
	\label{lem:1}
	A function $F(\cdot)$ = $\|\cdot\|_2^2$ is a $\sigma$-strongly convex ($\sigma \in (0,2]$) with respect to itself, that is, for $\forall \alpha \in [0,1]$ and $\bm{u},\bm{v} \in \mathbb{R}^d$, we have 
	\begin{align}
	F(\alpha\bm{u} + (1-\alpha)\bm{v}) \leq \alpha F(\bm{u}) + (1-\alpha) F(\bm{v}) \nonumber \\ -\frac{\sigma}{2} \alpha(1-\alpha)\|\bm{u} - \bm{v}\|_2^2
	\end{align}
	\begin{proof}
		\begin{align}
		F(&\alpha\bm{u} + (1-\alpha)\bm{v}) - \alpha F(\bm{u}) - (1-\alpha) F(\bm{v}) \nonumber \\  
		&= \|\alpha\bm{u} + (1-\alpha)\bm{v}\|_2^2 -\alpha\|\bm{u}\|_2^2 -(1-\alpha) \|\bm{v}\|_2^2 \nonumber \\
		&= -\alpha(1-\alpha)\|\bm{u}\|_2^2 + 2\alpha(1-\alpha)\langle \bm{u} , \bm{v} \rangle - \alpha(1-\alpha)\|\bm{v}\|_2^2 \nonumber \\
		&= -\alpha(1-\alpha)(\|\bm{u}\|_2^2-2\langle \bm{u} , \bm{v} \rangle+\|\bm{v}\|_2^2) \nonumber \\
		&= -\alpha(1-\alpha)\|\bm{u}-\bm{v}\|_2^2 \leq -\frac{\sigma}{2} \alpha(1-\alpha)\|\bm{u} - \bm{v}\|_2^2
		\end{align}
		The inequality follows by an obvious inequality $-\alpha(1-\alpha)\|\bm{u}-\bm{v}\|_2^2 \leq 0$. When $\sigma = 2$, the equality in this Lemma holds.  
	\end{proof}
\end{Lem}

The Rademacher complexity is one measure of the number of a family $\mathscr{F}$ of functions. Now we use convex duality to show that the Rademacher complexity is bounded by the number of inputs as follows.
\begin{Thm}
	\label{thm:3}
	Define $\mathscr{H} = \{\bm{h}: \|\bm{h}\|_2 \leq Z\}$ and $\mathscr{B} = \{\bm{w} \in S: \|\bm{w}\|_2^2 \leq B^2\}$, where $S \in \mathbb{R}^M$ is a subset of the dual space to the enhanced layer output vector space. The $\|\cdot\|_2^2$ is 2-strongly convex on $\mathbb{R}^M$ with respect to itself. Then, the Rademacher complexity is bounded by 
	\begin{align}
	\mathscr{R}_M(\mathscr{F}) \leq ZB\sqrt{\frac{1}{M}}
	\end{align}
	\begin{proof}
		Given a set of $\{\bm{h}_i\}$ $( i = 1, \dots, M)$ such that $\|\bm{h}_i\|_2 \leq Z$. Define $\bm{\theta} = \frac{1}{M} \sum_{i} \epsilon_i \bm{h}_i$, where $\epsilon_i, i = 1, \dots, M$ are Rademacher random variables defined above such that $\mathbb{E}[\epsilon_i] = 0$ and $\mathbb{E}[\epsilon_i^2] = 1$.
		
		Using Fenchel-Young inequality, we have
		\begin{align}
		\langle \bm{w} , \kappa\bm{\theta} \rangle \leq F(\bm{w}) + F^*(\bm{w})
		\end{align} 
		where $\kappa$ is an arbitrary positive real, and $F^*(\cdot)$ is the Fenchel conjugate of $F(\cdot)$. 
		
		Since the dual norm of the $l_2$ norm is itself, so we further have 
		\begin{align}
		\langle \bm{w} , \bm{\theta} \rangle \leq \frac{\|\bm{w}\|_2^2}{\kappa} + \frac{\|\kappa\bm{\bm{\theta}}\|_2^2}{\kappa}
		\end{align} 
		
		Following the condition of this theorem, we have 
		\begin{align}
		\sup_{\bm{w} \in \mathscr{B}} \langle \bm{w} , \bm{\theta} \rangle \leq \frac{B^2}{\kappa} + \frac{\|\kappa\bm{\bm{\theta}}\|_2^2}{\kappa}
		\end{align}
		
		We calculate expectation over $\epsilon_i$ on both sides and obtain 
		
		\begin{align}
		\label{eq:35}
		\mathbb{E}[\sup_{\bm{w} \in \mathscr{B}} \langle \bm{w} , \bm{\theta} \rangle] \leq \frac{B^2}{\kappa} + \frac{\mathbb{E}[\|\kappa\bm{\bm{\theta}}\|_2^2]}{\kappa}
		\end{align}
		
		We have proven $\|\cdot\|_2^2$ is a $\sigma$-strongly convex in Lemma \ref{lem:1}. Therefore, we can use the result of Lemma 2 in \cite{kakade2009complexity}. As a result, we can achieve the upper bound of $\mathbb{E}[\|\kappa\bm{\bm{\theta}}\|_2^2]$ as
		\begin{align}
		\label{eq:36}
		\mathbb{E}[\|\kappa\bm{\bm{\theta}}\|_2^2] \leq \frac{\kappa^2Z^2}{4M}
		\end{align}
		
		Substituting (\ref{eq:36}) into (\ref{eq:35}), we have 
		\begin{align}
		\label{eq:37}
		\mathbb{E}[\sup_{\bm{w} \in \mathscr{B}} \langle \bm{w} , \bm{\theta} \rangle] \leq \frac{B^2}{\kappa} + \frac{\kappa Z^2}{4M}
		\end{align} 
		
		We set $\kappa$ as $\sqrt{\frac{4MB^2}{Z^2}}$, and substitute it into (\ref{eq:37})
		\begin{align}
		\mathbb{E}[\sup_{\bm{w} \in \mathscr{B}} \langle \bm{w} , \bm{\theta} \rangle] \leq ZB\sqrt{\frac{1}{M}}
		\end{align}
		which is the desired result.
	\end{proof}  
\end{Thm}

Now we can bound the generalization error using the proved Rademacher complexity as follows. 
\begin{Thm}
	\label{thm:4}
	For the Lipschitz continuous loss function $\ell$ with Lipschitz constant $K$ in the RVFL+, with probability at least $1-\delta$ ($\delta \in (0,1)$) over the inputs with length $M$, for all $f \in \mathscr{F}$, the generalization risk minimization of the RVFL+ is bounded by 
	\begin{align}
	\mathfrak{L}(f) \leq \hat{\mathfrak{L}}(f) + 2KZB\sqrt{\frac{1}{M}}+KZB\sqrt{\frac{ln(1/\delta)}{2M}}
	\end{align}
	\begin{proof}
		We combine Theorem \ref{thm:2} and \ref{thm:3}, and set the bounded constant $c = KZB$. Then, we can obtain the desired result. 
	\end{proof}  
\end{Thm}

\section{Experiments and Results}
\label{sec:5}
In this section, we conduct several experiments to evaluate the proposed RVFL+ and KRVFL+ on 14 real-world datasets, including 1 binary classification dataset, 8 multi-class classification datasets and 5 regression datasets. Ten trials for each method are carried out, and the average results are reported. All simulations are carried out in a Matlab 2015b environment running in a PC machine with an Inter(R) Core(TM) i7-6700HD 2.60 GHZ CPU and 32 GB of RAM. 

\subsection{Evaluation on MNIST+} 
\subsubsection{Dataset}
The MNIST+ dataset is a popular testbed used in \cite{vapnik2009new,pechyony2010smo,li2016fast} for verifying the performance of the algorithms based on the LUPI. The MNIST+ is a handwritten digit recognition dataset, consisting of the images of the digits '5' and '8'. There are 100 training samples, 4,000 validation samples and 1,866 test samples in MNIST+. The original 28-by-28 gray-scale images from MNIST \cite{lecun1998gradient} are resized into 10-by-10 gray-scale images in order to increase challenges. In MNIST+, each sample in the training set and test set contains 100-dimensional attributes used as the normal features. Moreover, the additional privileged information in MNIST+ is 21-dimensional texture features based data, which is a holistic description of each image. We refer interested readers to \cite{vapnik2009new} for details.    
\subsubsection{Parameters selection}
For the RVFL+, we determine empirically the hyper-parameters $C$, $\gamma$, $u$ and $P$, as well as the nonlinear activation function on the validation set. First of all, we select the nonlinear activation function. Table \ref{table:1} reports the performance of the RVFL+ with different activation functions in terms of accuracy. As shown by Table \ref{table:1}, the RVFL+ with the triangular basis function outperforms others, and thus, we use the triangular basis function as the activation function in this case. However, we found that there does not exist a general rule for choosing the activation function. Therefore, we need to determine the activation function in the RVFL+ for different tasks.           
\begin{table}[h]
	\renewcommand{\arraystretch}{1.3}
	\caption{The evaluation of the RVFL+ with different activation functions}
	\label{table:1}
	\centering
	\begin{tabular}{|c|c|c|}
		\hline
		Method & Activation Function & Accuracy ($\%$) \\
		\hline
		\multirow{5}{*}{RVFL+}  &Sigmoid function & 85.22 $\pm$ 0.38\\
		\cline{2-3}
		& Sine function& 85.06 $\pm$ 1.06 \\
		\cline{2-3}
		& Hardlim function& 87.01 $\pm$ 1.56 \\
		\cline{2-3}
		& Triangular basis function & $\bm{90.41}\pm \bm{0.74}$\\
		\cline{2-3}
		& Radial basis function & 84.36 $\pm$ 0.54\\
		\hline
	\end{tabular}
\end{table} 

\begin{figure}[ht]
	\centering
	\includegraphics[width=0.7\textwidth]{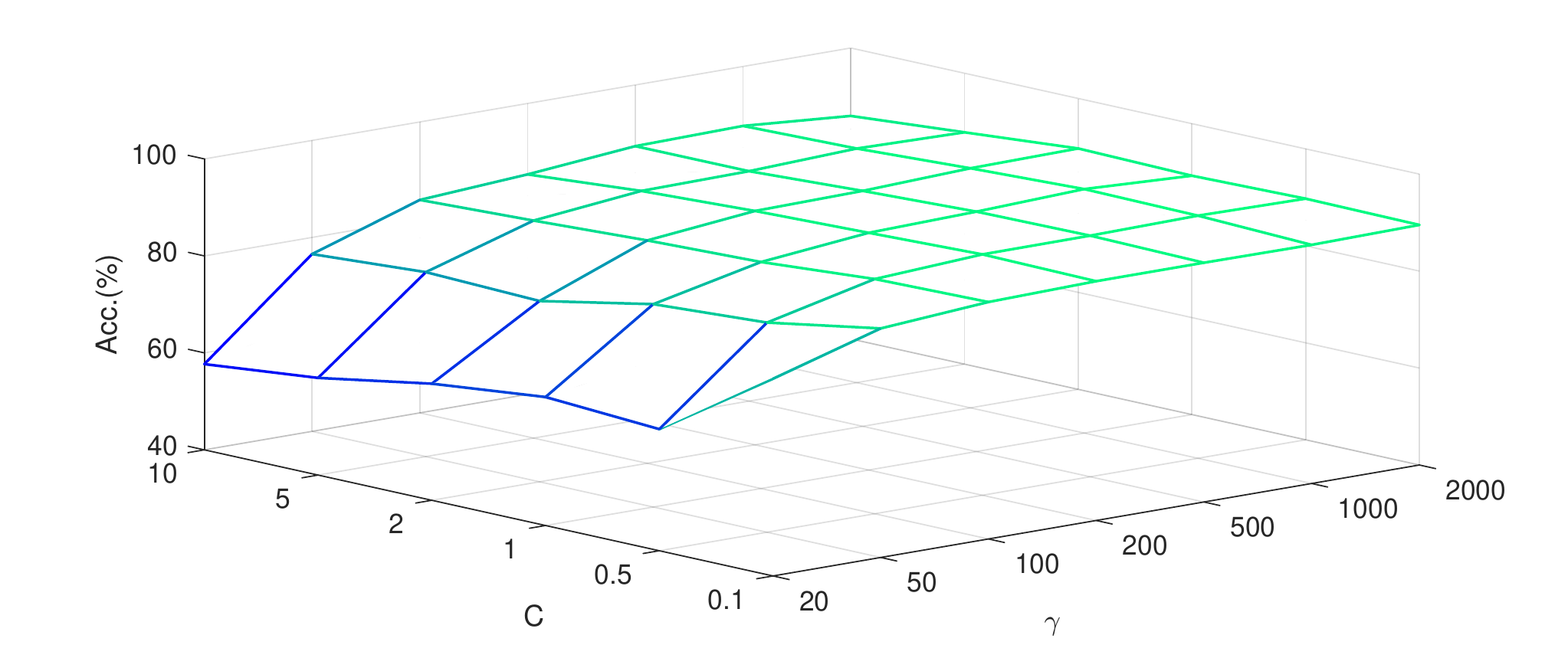}
	\caption{The performance of RVFL+ with different user-defined parameters $C$ and $\gamma$, in which the RVFL+ uses the triangular basis function as the activation function.}
	\label{fig:2}
\end{figure}

The user-defined parameters $C$ and $\gamma$ are chosen through a random search \cite{bergstra2012random} within $[10^{-5}, 10^5]$, and $u$ is also selected from the interval $[2^{-5}, 2^{-4.5}, \dots, 2^{4.5}, 2^5]$  in the same fashion. In Figure \ref{fig:2}, we only show part of results of selecting the parameters $C$ and $\gamma$, and ignore numerous ones having the worse performance. As shown in Figure \ref{fig:2}, when $C$ and $\gamma$ are set as 1 and 1,000 respectively, the RVFL+ can achieve the best performance. Therefore, we determine $C$ and $\gamma$ as 1 and 1,000 in the following comparison. Additionally, when the number of enhancement nodes $P$ is sufficient large, the RVFL+ can achieve good generalization performance. Therefore balancing the effectiveness and efficiency, $P$ is determined as 1,000 in all experiments. Furthermore, we found that a 'suitable' $u$ can improve around $2\%$ accuracy for the RVFL+, and thus, we empirically determine an optimal $u$ in different tasks. In this experiment, the positive factor $u$ is empirically set as $2^{2.5}$. Due to space limitations, we omit these procedures of selecting $P$ and $u$.

\begin{figure*}[ht!]
	\centering
	\subfloat[The best performance of KRVFL+ corresponding to different $\gamma$.]{\includegraphics[width = 0.65\textwidth]{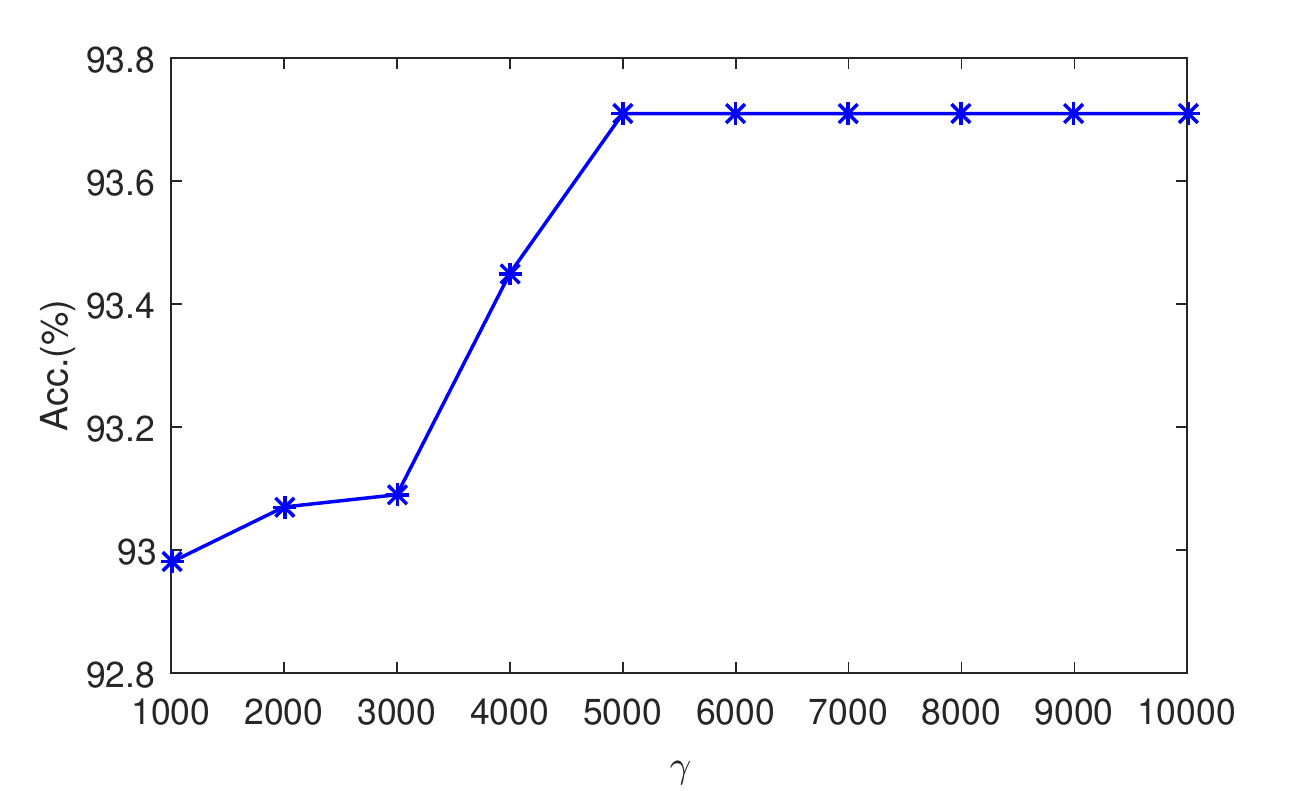}%
		\label{fig:4}}
	\hfil
	\subfloat[The performance of KRVFL+ with different user-defined parameters $C$ and $\tau$, when $\gamma = 5,000$.]{\includegraphics[width = 0.65\textwidth]{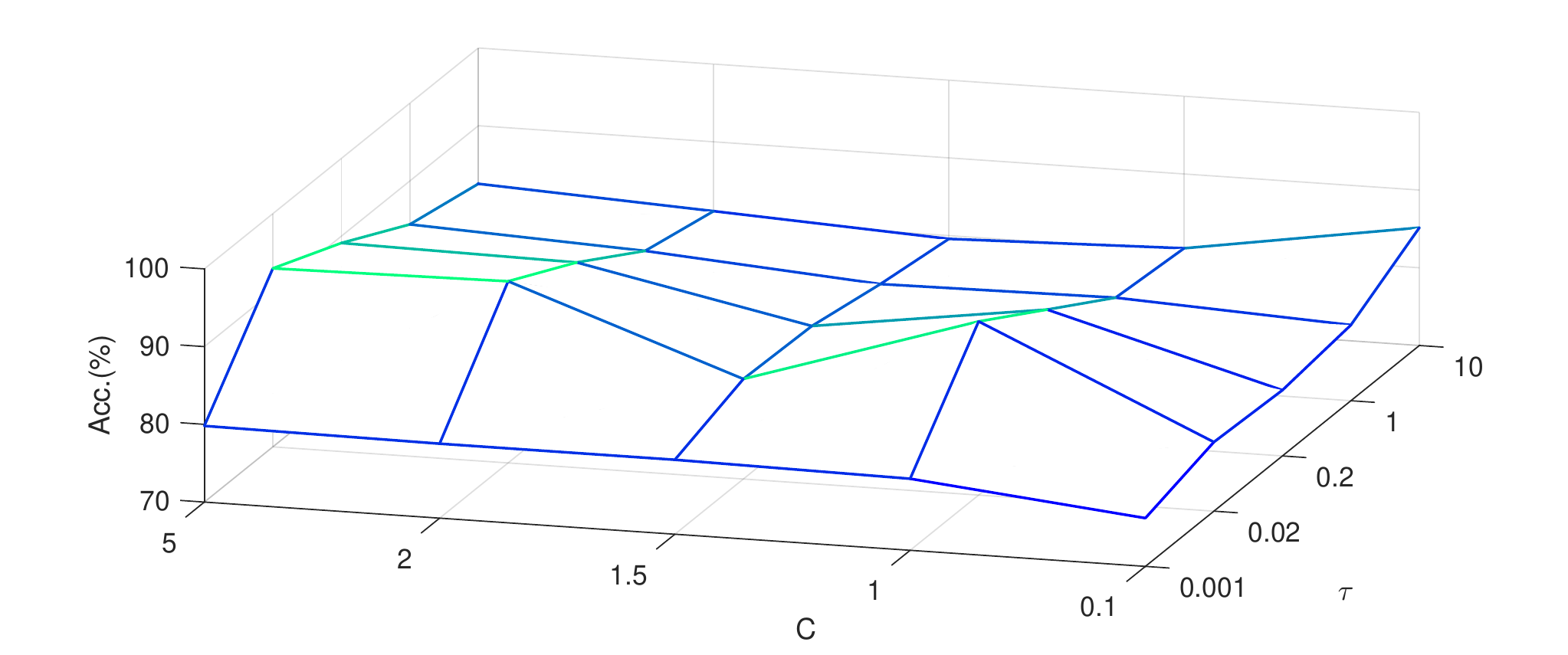}%
		\label{fig:3}}
	\caption{Parameter selection for KRVFL+.}
\end{figure*}

There are two kernel functions in the KRVFL+. In all simulations, we use the commonly-used Gaussian kernel function as the general Mercer kernel, while the other kernel function is defined as a linear kernel function. The user-specified kernel parameter $\tau$, $C$ as well as $\gamma$ are determined on the validation set. Seen from Figure \ref{fig:4}, when the $\gamma$ is sufficient large (more than 5,000), the KRVFL+ is insensitive to this parameter. Therefore, we set $\gamma$ as 5,000 for KRVFL+ in all experiments. Figure \ref{fig:3} illustrates the performance of KRVFL+ with different parameters $C$ and $\tau$, when $\gamma = 5,000$. We found when $C$ and $\tau$ are in the interval $[1,2]$ and $[0.02,0.03]$ respectively, the RVFL+ can achieve the best performance in this case.  

For other approaches, due to space limitations, we omit the procedure of selecting the hyper-parameters. We empirically determine these user-specified parameters on the validation set, and pick the one having the best performance.

\subsubsection{Experimental Results and Discussion}

We compare with state-of-the-art approaches on MNIST+ in terms of accuracy ($\%$) and time (s), including SVM \cite{CC01a}, gSMO-SVM+ \cite{pechyony2010smo}, CVX-SVM+ \cite{liang2008connection}, MAT-SVM+ \cite{li2016fast}, Fast SVM+ \cite{li2016fast} and RVFL \cite{pao1994learning}. The experimental results are shown in Table \ref{table:2}.
\begin{table}[h]
	\renewcommand{\arraystretch}{1.3}
	\caption{The comparisons with state-of-the-art approaches on MNIST+ in terms of accuracy ($\%$) and time (s)}
	\label{table:2}
	\centering
	\begin{tabular}{|c|c|c|c|}
		\hline
		Methods & Kernel/Activation Function & Acc. ($\%$)& time (s) \\
		\hline
		SVM \cite{CC01a} & Linear kernel &	81.73 & 0.009 \\
		\hline
		gSMO-SVM+ \cite{pechyony2010smo}  & Linear kernel& 84.35 & 1.103 \\
		\hline
		Fast SVM+ \cite{li2016fast} & Linear kernel & 84.62 & 0.146 \\
		\hline
		SVM \cite{CC01a} & Gaussian kernel & 92.34 & 0.009\\
		\hline
		gSMO-SVM+ \cite{pechyony2010smo}  &Gaussian kernel & 92.77 & 1.206 \\
		\hline
		CVX-SVM+ \cite{liang2008connection} &Gaussian kernel & 93.14 & 11.746 \\
		\hline
		MAT-SVM+ \cite{li2016fast} &Gaussian kernel & 93.14 & 0.572 \\
		\hline
		Fast SVM+ \cite{li2016fast} &Gaussian kernel & 93.15 & 0.039 \\
		\hline
		RVFL \cite{pao1994learning}&Triangular basis & 87.08 & 0.021\\
		\hline
		RVFL+ &Triangular basis& 90.40 & 0.027\\
		\hline
		KRVFL+ &Gaussian kernel& $\bm{93.71}$ & 0.005\\
		\hline
	\end{tabular}
\end{table} 

From Table \ref{table:2}, we see that the KRVFL+ has the best performance in terms of effectiveness and efficiency. Although the proposed RVFL+ is slightly worse than the kernel-based SVM+, this approach is much better than all linear SVM+. In contrast to SVM and RVFL, this experimental result illustrates that benefits of the LUPI-based approaches accrue with little additional computational cost.

\subsection{Evaluation on Classification and Regression Datasets}

In this section, we firstly compare with Fast SVM+ \cite{li2016fast}, MAT-SVM+ \cite{li2016fast}, SVM \cite{CC01a} and RVFL \cite{pao1994learning} on 8 real-world multi-class classification datasets from UCI machine learning repository \cite{Lichman:2013}, which cover a large range of multi-class classification tasks. The previous work \cite{li2016fast} has verified that the Fast SVM+ outperforms other state-of-the-art LUPI-based algorithms. To avoid duplication, these algorithms are not included for comparison. The statistics of the UCI classification datasets are illustrated in Table \ref{table:3}, including the number of training and test data, attributes, the number of the normal features, the number of the privileged features and classes. We split attributes of each classification dataset mentioned above in half. We select one part as normal ones and the others as privileged. We use $L_1$ normalization to pre-process all samples.  We use two-fold cross validation (CV) for Shuttle, five-fold CV for Abalone/Red Wine Quality/White Wine Quality, and ten-fold CV for the rest of the datasets. We found that both RVFL+ and RVFL can achieve the best performance, when using the sigmoid function as the nonlinear activation function in this experiment. Due to space limitations, we only report the final experimental results of all comparisons, and omit the procedure of selecting hyper-parameters and pick the one having the best performance. The experimental results are reported in Table \ref{table:4}.

\begin{table}[h]
	\renewcommand{\arraystretch}{1.3}
	\caption{The statistics of UCI classification datasets}
	\label{table:3}
	\centering
	\resizebox{\textwidth}{25mm}{
	\begin{tabular}{|c|c|c|c|c|c|c|}
		\hline
		Problems & Training Data & Test Data & Attributes & Normal features size & Privileged features size & Classes \\
		\hline
		Iris &	90 & 60 & 4 & 2 & 2 & 3 \\
		\hline
		Glass & 140 & 74 & 9 & 5 & 4 &5 \\
		\hline
		Abalone & 2,000 & 2,177 & 8 & 4 & 4 &3 \\
		\hline
		Wine & 100 & 78 & 13& 7 & 6 &3\\
		\hline
		Red Wine Quality & 1,000 & 599 & 11 & 6 & 5 & 3\\
		\hline
		White Wine Quality & 3,000 & 1,898 & 11 &  6&  5 & 7 \\
		\hline
		Shuttle & 43,500 & 14,500 & 9 & 5 & 4 &7 \\
		\hline
		Segment & 100 & 110 & 19 & 11 & 8 & 7 \\
		\hline
	\end{tabular}}
\end{table}

\begin{table}[h]
	\renewcommand{\arraystretch}{1.3}
	\caption{The comparisons with state-of-the-art approaches on classification datasets with half of features in terms of accuracy ($\%$) and time (s)}
	\label{table:4}
	\centering
	\resizebox{\textwidth}{25mm}{
	\begin{tabular}{|c|c|c|c|c|c|c|c|c|c|c|c|c|}
		\hline
		\multirow{2}{*}{Problems} & \multicolumn{2}{c|}{KRVFL+} & \multicolumn{2}{c|}{RVFL+} & \multicolumn{2}{c|}{RVFL \cite{pao1994learning}}  & \multicolumn{2}{c|}{Fast SVM+ \cite{li2016fast}}  & \multicolumn{2}{c|}{MAT-SVM+ \cite{li2016fast}}  & \multicolumn{2}{c|}{SVM \cite{CC01a}} \\
		\cline{2-13}
		& Acc. ($\%$) & time (s) & Acc. ($\%$) & time (s) & Acc. ($\%$) & time (s) & Acc. ($\%$) & time (s) & Acc. ($\%$) & time (s) & Acc. ($\%$) & time (s) \\
		\hline
		Iris &	$\bm{77.25}$ & 0.001  & 74.38 & 0.029 & 69.87 & 0.025 & 76.85 & 0.002  & 76.27 & 0.353 & 70.00 & 0.001 \\
		\hline
		Glass &	$\bm{58.27}$ & 0.001  & 57.93 & 0.062 & 51.85 & 0.057 & 57.58 & 0.003  & 57.69 & 0.092 & 53.48 & 0.001 \\
		\hline
		Abalone & 64.52	 & 0.913  & $\bm{66.89}$ & 0.268 & 53.79 & 0.247 & 65.82 & 0.518  & 65.91 & 16.48 & 54.78 & 0.055\\
		\hline
		Wine & $\bm{94.37}$	 & 0.002  & 89.27 & 0.108  & 84.28 & 0.091 & 93.47 & 0.006  & 93.51 & 0.105  & 83.30 & 0.002\\
		\hline
		Red Wine Quality &	65.54 &  0.279 & $\bm{66.17}$ &0.157  & 58.57 & 0.143 & 65.57 & 0.101 & 65.48 & 168.2 & 56.56 & 0.041\\
		\hline
		White Wine Quality & $\bm{57.92}$ & 0.216  & 57.11 & 0.329 & 50.12 & 0.286 & 57.01 &  1.382 & 57.02 & $>$3600 & 49.89 & 0.398\\
		\hline
		Shuttle & $\bm{99.52}$	 & 436.1  & 99.26 & 2.431 & 98.56  & 2.378 & 99.17 & 415.2  & 99.38 & $>$3600 & 98.31 & 285.3 \\
		\hline
		Segment & 29.47	 & 0.297  & 30.01 & 0.692 & 20.16  & 0.653 &  $\bm{30.58}$ & 0.003  & 28.57 & 0.201 & 19.04 & 0.001 \\
		\hline
	\end{tabular}}
\end{table}

Seen from Table \ref{table:4}, except for the Abalone, Red Wine Quality and Segment, the KRVFL+ can achieve the best performance in terms of accuracy among all comparisons. While the RVFL+ outperforms other comparisons including KRVFL+ on the Abalone and Red Wine Quality. Likewise to the first experiment in this paper, the performance of methods with the LUPI paradigm including KRVFL+, RVFL+, Fast SVM+ and MAT-SVM+ are better remarkably than both SVM and RVFL. It means that the LUPI paradigm can significantly enhance the performance in terms of effectiveness. 

Furthermore, we compare with SVR \cite{CC01a} and RVFL \cite{pao1994learning} on 5 real-world regression datasets, which cover different categories from \cite{spyromitros2016multi}. The statistics of these regression datasets are summarized in Table \ref{table:5}.  We use $L_1$ normalization to pre-process all data. We use two-fold cross validation (CV) for SCM1D/SCM20D, ten-fold CV for the rest of the datasets. We also found that both RVFL+ and RVFL can achieve the best performance, when using the sigmoid function as the nonlinear activation function in this experiment. We also omit the procedure of selecting user-specified parameters. The performance is measured by the commonly-used Root Mean Square Error (RMSE). The experimental results are shown in Table \ref{table:6}. The experimental results illustrate the advantage of the new learning paradigm. Notice that both KRVFL+ and RVFL+ outperform RVFL and SVR on all regression datasets.  
\begin{table}[h]
	\renewcommand{\arraystretch}{1.3}
	\caption{The statistics of regression datasets}
	\label{table:5}
	\centering
	\resizebox{\textwidth}{20mm}{
	\begin{tabular}{|c|c|c|c|c|c|}
		\hline
		Problems & Training Data & Test Data & Attributes & Normal features size & Privileged features size  \\
		\hline
		ANDRO &	23 & 26 & 30 & 15 & 15 \\
		\hline
		EDM & 140 & 74 & 16 & 8 & 8\\
		\hline
		SLUMP & 53 & 50 & 7 & 4 & 3 \\
		\hline
		SCM1D & 5,000 & 4,803 & 280 & 140 & 140\\
		\hline
		SCM20D & 6,000 & 2,966 & 61 & 31 & 30\\
		\hline
	\end{tabular}}
\end{table}

\begin{table}[h]
	\renewcommand{\arraystretch}{1.3}
	\caption{The comparisons with state-of-the-art approaches on regression datasets with half of features in terms of RMSE and time (s)}
	\label{table:6}
	\centering
	\resizebox{\textwidth}{20mm}{
	\begin{tabular}{|c|c|c|c|c|c|c|c|c|}
		\hline
		\multirow{2}{*}{Problems} & \multicolumn{2}{c|}{KRVFL+} & \multicolumn{2}{c|}{RVFL+} & \multicolumn{2}{c|}{RVFL \cite{pao1994learning}}  & \multicolumn{2}{c|}{SVR \cite{CC01a}} \\
		\cline{2-9}
		& RMSE & time (s) & RMSE & time (s) & RMSE & time (s) & RMSE & time (s)  \\
		\hline
		ANDRO &	$\bm{3.516}$  & 0.0001  & 3.973 & 0.0452 & 4.967 & 0.0416 &  4.422 & 0.0225    \\ 
		\hline
		EDM & 0.057 & 0.0005 & $\bm{0.052}$ & 0.0319 & 0.061 & 0.0283 & 0.069 & 0.0004 \\
		\hline
		SLUMP & $\bm{34.928}$ & 0.0002 & 36.449 & 0.0423 & 39.358 & 0.0317 & 40.486 & 0.0006 \\
		\hline
		SCM1D & 60.237 & 1.8542 & $\bm{59.685}$ & 9.3154 & 65.917 & 9.0365 &  63.872 & 5.9286\\ 
		\hline
		SCM20D & $\bm{55.283}$ & 3.576 & 57.268 & 2.5028 & 60.843 & 2.012 & 61.815 & 2.451  \\
		\hline		
	\end{tabular}}
\end{table} 

\subsection{Evaluation on Noise Datasets}

In this section, we empirically evaluate whether or not the LUPI paradigm can significantly enhance the performance in terms of effectiveness on noise datasets. We also select the same classification and regression datasets mentioned above. We add white noise with 10 dBW to both training and test samples on both classification and regression datasets. While the original training data without white noise is used as privileged information for all LUPI-based methods. In addition, the experimental setup is the same as that of the last experiment.    

Seen from Table \ref{table:7}, except for the White Wine Quality and the Red Wine Quality, the KRVFL+ can achieve the best performance in terms of accuracy among all comparisons. While the RVFL+ outperforms other comparisons including KRVFL+ on the White Wine Quality. Likewise to the above experiment, the performance of these methods with the LUPI paradigm including KRVFL+, RVFL+, Fast SVM+ and MAT-SVM+ are better remarkably than both SVM and RVFL.
\begin{table}[h]
	\renewcommand{\arraystretch}{1.3}
	\caption{The comparisons with state-of-the-art approaches on noise classification datasets in terms of accuracy ($\%$) and time (s)}
	\label{table:7}
	\centering
	\resizebox{\textwidth}{25mm}{
		\begin{tabular}{|c|c|c|c|c|c|c|c|c|c|c|c|c|}
			\hline
			\multirow{2}{*}{Problems} & \multicolumn{2}{c|}{KRVFL+} & \multicolumn{2}{c|}{RVFL+} & \multicolumn{2}{c|}{RVFL \cite{pao1994learning}}  & \multicolumn{2}{c|}{Fast SVM+ \cite{li2016fast}}  & \multicolumn{2}{c|}{MAT-SVM+ \cite{li2016fast}}  & \multicolumn{2}{c|}{SVM \cite{CC01a}} \\
			\cline{2-13}
			& Acc. ($\%$) & time (s) & Acc. ($\%$) & time (s) & Acc. ($\%$) & time (s) & Acc. ($\%$) & time (s) & Acc. ($\%$) & time (s) & Acc. ($\%$) & time (s) \\
			\hline
			Iris & $\bm{51.67}$	 & 0.002  & 46.67 & 0.093 & 43.33  & 0.078  & 49.63 & 0.007  & 49.62 & 1.102  & 43.57 & 0.002 \\
			\hline
			Glass & $\bm{39.19}$ & 0.002 & 36.49 & 0.154 & 29.03 & 0.141 & 38.95 & 0.01 & 38.95 & 0.231 & 29.75& 0.002\\
			\hline
			Abalone & $\bm{43.50}$ & 0.219 & 42.77 & 0.643 & 42.31 & 0.578 & 42.90 & 1.194 & 42.87 & 37.13 &42.09 & 0.134\\
			\hline
			Wine & $\bm{83.00}$ & 0.003 & 74.36 & 0.259 & 61.54 & 0.234 & 82.05 & 0.016 & 81.98 & 0.250 & 61.77& 0.004 \\
			\hline
			Red Wine Quality & 53.92 & 0.672 &  52.59 & 0.392 & 52.09 & 0.343 & $\bm{53.98}$ & 0.307 & 53.91 & 378.3 & 52.14& 0.082 \\
			\hline
			White Wine Quality & 47.74 & 0.514 & $\bm{49.00}$ & 0.824 & 45.47& 0.719 & 47.13 & 3.592 & 47.02 & $>$ 3600 & 45.86& 1.034 \\
			\hline
			Shuttle & $\bm{98.48}$ & 1087.6 & 97.18 & 5.982 & 96.56 & 5.375 & 98.33 & 976.4 & 98.21 & $>$ 3600 & 96.62& 713.2 \\
			\hline
			Segment & $\bm{79.09}$ & 0.743 & 77.27 & 1.792 & 73.64& 1.676 & 78.52& 0.008 & 78.41 & 0.491 & 73.38 & 0.003 \\
			\hline
		\end{tabular}}
\end{table}

Additionally, the regression experimental results are shown in Table \ref{table:8}. The experimental results illustrate the advantage of the new learning paradigm. Although the RVFL+ and KRVFL+ achieve sightly better performance than the other two on EDM and SLUMP due to the small size of these two datasets, the two proposed models have distinct advantage on the big size of datasets such as SCM1D and SCM20D. 

\begin{table}[h]
	\renewcommand{\arraystretch}{1.3}
	\caption{The comparisons with state-of-the-art approaches on noise regression datasets in terms of rmse and time (s)}
	\label{table:8}
	\centering
	\resizebox{\textwidth}{20mm}{
		\begin{tabular}{|c|c|c|c|c|c|c|c|c|}
			\hline
			\multirow{2}{*}{Problems} & \multicolumn{2}{c|}{KRVFL+} & \multicolumn{2}{c|}{RVFL+} & \multicolumn{2}{c|}{RVFL \cite{pao1994learning}}  & \multicolumn{2}{c|}{SVR \cite{CC01a}} \\
			\cline{2-9}
			& RMSE & time (s) & RMSE & time (s) & RMSE & time (s) & RMSE & time (s)  \\
			\hline
			ANDRO &	$\bm{2.097}$  & 0.0004  & 2.151 & 0.1125 & 2.891 & 0.1094 & 2.853 & 0.0509    \\ 
			\hline
			EDM & 0.178 & 0.0014 & $\bm{0.175}$ & 0.0836 & 0.182 & 0.0781 & 0.179 & 0.0012 \\
			\hline
			SLUMP & 3.503 & 0.0005 & $\bm{3.501}$ & 0.1091 & 3.508 & 0.0781 & 3.512 & 0.0012 \\
			\hline
			SCM1D & $\bm{45.816}$ & 5.1495 & 49.007 & 22.328 & 51.618  & 23.755 & 51.962 & 12.723\\ 
			\hline
			SCM20D & $\bm{99.47}$ & 7.958 & 103.08 & 5.347 & 109.91 & 4.9844 & 107.86 & 5.377  \\
			\hline		
	\end{tabular}}
\end{table}

\section{Conclusion}
\label{sec:6}
In this paper, we first present a LUPI based random vector functional link network called RVFL+, which combines randomized neural networks with the new LUPI paradigm. The newly-derived LUPI-based randomized neural networks leverage the benefits of both. More significantly, the RVFL+ offers a new learning mechanism for the RVFL network, in which the additional privileged information is considered as a teacher during the training stage. Therefore, we note that the working mechanism of the newly-derived RVFL+ is in analogy to $\emph{Teacher-Student Interaction}$ \cite{vapnik2015learning} in human learning process. As a result, the RVFL+ can achieve better generalization performance than the RVFL in real-world tasks. In addition, from an optimization perspective, the RVFL+ has milder and simpler optimization constraints than the SVM+ \cite{vapnik2009new}, which results that the RVFL+ can obtain a closed-form solution. Moreover, the RVFL+ can perform in conjunction with the kernel trick, which is defined as KRVFL+. The novel KRVFL+ is powerful enough to handle the more complicated nonlinear relationships between high-dimensional input data. Furthermore, we explore the statistical property of the proposed RVFL+, and establish a tight generalization error bound based on the Rademacher complexity for the RVFL+. Competitive experimental results on 14 diverse real-world datasets confirm the efficiency and effectiveness of the new RVFL+ and KRVFL+, which can achieve better generalization performance than the state-of-the-art methods.

\section*{Acknowledgment}

The authors would like to thank Prof. C. L. Philip Chen from University of Macau for valuable discussions on the random vector functional link network and the kernel ridge regression. This work was
supported in part by the Science and Technology Development Fund of Macao S.A.R (FDCT) under MoST-FDCT Joint Grant 015/2015/AMJ and FDCT grant 121/2016/A3, 194/2017/A3; in part by University of Macau under grant MYRG2016-00160-FST.



\bibliographystyle{elsarticle-num-names} 
\bibliography{references}

\begin{thebibliography}{58}
\providecommand{\natexlab}[1]{#1}
\providecommand{\url}[1]{\texttt{#1}}
\providecommand{\urlprefix}{URL }
\expandafter\ifx\csname urlstyle\endcsname\relax
  \providecommand{\doi}[1]{doi:\discretionary{}{}{}#1}\else
  \providecommand{\doi}[1]{doi:\discretionary{}{}{}\begingroup
  \urlstyle{rm}\url{#1}\endgroup}\fi
\providecommand{\bibinfo}[2]{#2}

\bibitem[{Vapnik and Vashist(2009)}]{vapnik2009new}
\bibinfo{author}{V.~Vapnik}, \bibinfo{author}{A.~Vashist}, \bibinfo{title}{A
  new learning paradigm: Learning using privileged information},
  \bibinfo{journal}{Neural networks} \bibinfo{volume}{22}~(\bibinfo{number}{5})
  (\bibinfo{year}{2009}) \bibinfo{pages}{544--557}.

\bibitem[{Pechyony and Vapnik(2010)}]{pechyony2010theory}
\bibinfo{author}{D.~Pechyony}, \bibinfo{author}{V.~Vapnik}, \bibinfo{title}{On
  the theory of learnining with privileged information}, in:
  \bibinfo{booktitle}{Advances in neural information processing systems},
  \bibinfo{pages}{1894--1902}, \bibinfo{year}{2010}.

\bibitem[{Vapnik and Izmailov(2015)}]{vapnik2015learning}
\bibinfo{author}{V.~Vapnik}, \bibinfo{author}{R.~Izmailov},
  \bibinfo{title}{Learning using privileged information: similarity control and
  knowledge transfer.}, \bibinfo{journal}{Journal of machine learning research}
  \bibinfo{volume}{16}~(\bibinfo{number}{20232049}) (\bibinfo{year}{2015})
  \bibinfo{pages}{55}.

\bibitem[{Grant et~al.(2008)Grant, Boyd, and Ye}]{grant2008cvx}
\bibinfo{author}{M.~Grant}, \bibinfo{author}{S.~Boyd}, \bibinfo{author}{Y.~Ye},
  \bibinfo{title}{CVX: Matlab software for disciplined convex programming},
  \bibinfo{year}{2008}.

\bibitem[{Pechyony et~al.(2010)Pechyony, Izmailov, Vashist, and
  Vapnik}]{pechyony2010smo}
\bibinfo{author}{D.~Pechyony}, \bibinfo{author}{R.~Izmailov},
  \bibinfo{author}{A.~Vashist}, \bibinfo{author}{V.~Vapnik},
  \bibinfo{title}{SMO-Style Algorithms for Learning Using Privileged
  Information.}, in: \bibinfo{booktitle}{DMIN}, \bibinfo{pages}{235--241},
  \bibinfo{year}{2010}.

\bibitem[{Li et~al.(2016)Li, Dai, Tan, Xu, and Van~Gool}]{li2016fast}
\bibinfo{author}{W.~Li}, \bibinfo{author}{D.~Dai}, \bibinfo{author}{M.~Tan},
  \bibinfo{author}{D.~Xu}, \bibinfo{author}{L.~Van~Gool}, \bibinfo{title}{Fast
  algorithms for linear and kernel svm+}, in: \bibinfo{booktitle}{Proceedings
  of the IEEE Conference on Computer Vision and Pattern Recognition},
  \bibinfo{pages}{2258--2266}, \bibinfo{year}{2016}.

\bibitem[{Pechyony and Vapnik(2011)}]{pechyony2011fast}
\bibinfo{author}{D.~Pechyony}, \bibinfo{author}{V.~Vapnik},
  \bibinfo{title}{Fast optimization algorithms for solving SVM+},
  \bibinfo{journal}{Stat. Learning and Data Science} \bibinfo{volume}{1}.

\bibitem[{Feyereisl et~al.(2014)Feyereisl, Kwak, Son, and
  Han}]{feyereisl2014object}
\bibinfo{author}{J.~Feyereisl}, \bibinfo{author}{S.~Kwak},
  \bibinfo{author}{J.~Son}, \bibinfo{author}{B.~Han}, \bibinfo{title}{Object
  localization based on structural SVM using privileged information}, in:
  \bibinfo{booktitle}{Advances in Neural Information Processing Systems},
  \bibinfo{pages}{208--216}, \bibinfo{year}{2014}.

\bibitem[{Fouad et~al.(2013)Fouad, Tino, Raychaudhury, and
  Schneider}]{fouad2013incorporating}
\bibinfo{author}{S.~Fouad}, \bibinfo{author}{P.~Tino},
  \bibinfo{author}{S.~Raychaudhury}, \bibinfo{author}{P.~Schneider},
  \bibinfo{title}{Incorporating privileged information through metric
  learning}, \bibinfo{journal}{IEEE transactions on neural networks and
  learning systems} \bibinfo{volume}{24}~(\bibinfo{number}{7})
  (\bibinfo{year}{2013}) \bibinfo{pages}{1086--1098}.

\bibitem[{Xu et~al.(2015)Xu, Li, and Xu}]{xu2015distance}
\bibinfo{author}{X.~Xu}, \bibinfo{author}{W.~Li}, \bibinfo{author}{D.~Xu},
  \bibinfo{title}{Distance metric learning using privileged information for
  face verification and person re-identification}, \bibinfo{journal}{IEEE
  transactions on neural networks and learning systems}
  \bibinfo{volume}{26}~(\bibinfo{number}{12}) (\bibinfo{year}{2015})
  \bibinfo{pages}{3150--3162}.

\bibitem[{Lapin et~al.(2014)Lapin, Hein, and Schiele}]{lapin2014learning}
\bibinfo{author}{M.~Lapin}, \bibinfo{author}{M.~Hein},
  \bibinfo{author}{B.~Schiele}, \bibinfo{title}{Learning using privileged
  information: SVM+ and weighted SVM}, \bibinfo{journal}{Neural Networks}
  \bibinfo{volume}{53} (\bibinfo{year}{2014}) \bibinfo{pages}{95--108}.

\bibitem[{Sharmanska et~al.(2013)Sharmanska, Quadrianto, and
  Lampert}]{sharmanska2013learning}
\bibinfo{author}{V.~Sharmanska}, \bibinfo{author}{N.~Quadrianto},
  \bibinfo{author}{C.~H. Lampert}, \bibinfo{title}{Learning to rank using
  privileged information}, in: \bibinfo{booktitle}{Proceedings of the IEEE
  International Conference on Computer Vision}, \bibinfo{pages}{825--832},
  \bibinfo{year}{2013}.

\bibitem[{Bengio et~al.(2013)Bengio, Yao, Alain, and
  Vincent}]{bengio2013generalized}
\bibinfo{author}{Y.~Bengio}, \bibinfo{author}{L.~Yao},
  \bibinfo{author}{G.~Alain}, \bibinfo{author}{P.~Vincent},
  \bibinfo{title}{Generalized denoising auto-encoders as generative models},
  in: \bibinfo{booktitle}{Advances in Neural Information Processing Systems},
  \bibinfo{pages}{899--907}, \bibinfo{year}{2013}.

\bibitem[{Krizhevsky et~al.(2012)Krizhevsky, Sutskever, and
  Hinton}]{krizhevsky2012imagenet}
\bibinfo{author}{A.~Krizhevsky}, \bibinfo{author}{I.~Sutskever},
  \bibinfo{author}{G.~E. Hinton}, \bibinfo{title}{Imagenet classification with
  deep convolutional neural networks}, in: \bibinfo{booktitle}{Advances in
  neural information processing systems}, \bibinfo{pages}{1097--1105},
  \bibinfo{year}{2012}.

\bibitem[{Hinton et~al.(2006)Hinton, Osindero, and Teh}]{hinton2006fast}
\bibinfo{author}{G.~E. Hinton}, \bibinfo{author}{S.~Osindero},
  \bibinfo{author}{Y.-W. Teh}, \bibinfo{title}{A fast learning algorithm for
  deep belief nets}, \bibinfo{journal}{Neural computation}
  \bibinfo{volume}{18}~(\bibinfo{number}{7}) (\bibinfo{year}{2006})
  \bibinfo{pages}{1527--1554}.

\bibitem[{Hochreiter and Schmidhuber(1997)}]{hochreiter1997long}
\bibinfo{author}{S.~Hochreiter}, \bibinfo{author}{J.~Schmidhuber},
  \bibinfo{title}{Long short-term memory}, \bibinfo{journal}{Neural
  computation} \bibinfo{volume}{9}~(\bibinfo{number}{8}) (\bibinfo{year}{1997})
  \bibinfo{pages}{1735--1780}.

\bibitem[{Pao and Phillips(1995)}]{pao1995functional}
\bibinfo{author}{Y.-H. Pao}, \bibinfo{author}{S.~M. Phillips},
  \bibinfo{title}{The functional link net and learning optimal control},
  \bibinfo{journal}{Neurocomputing} \bibinfo{volume}{9}~(\bibinfo{number}{2})
  (\bibinfo{year}{1995}) \bibinfo{pages}{149--164}.

\bibitem[{Pao and Takefuji(1992)}]{pao1992functional}
\bibinfo{author}{Y.-H. Pao}, \bibinfo{author}{Y.~Takefuji},
  \bibinfo{title}{Functional-link net computing: theory, system architecture,
  and functionalities}, \bibinfo{journal}{Computer}
  \bibinfo{volume}{25}~(\bibinfo{number}{5}) (\bibinfo{year}{1992})
  \bibinfo{pages}{76--79}.

\bibitem[{Pao et~al.(1994)Pao, Park, and Sobajic}]{pao1994learning}
\bibinfo{author}{Y.-H. Pao}, \bibinfo{author}{G.-H. Park},
  \bibinfo{author}{D.~J. Sobajic}, \bibinfo{title}{Learning and generalization
  characteristics of the random vector functional-link net},
  \bibinfo{journal}{Neurocomputing} \bibinfo{volume}{6}~(\bibinfo{number}{2})
  (\bibinfo{year}{1994}) \bibinfo{pages}{163--180}.

\bibitem[{Igelnik and Pao(1995)}]{igelnik1995stochastic}
\bibinfo{author}{B.~Igelnik}, \bibinfo{author}{Y.-H. Pao},
  \bibinfo{title}{Stochastic choice of basis functions in adaptive function
  approximation and the functional-link net}, \bibinfo{journal}{IEEE
  Transactions on Neural Networks} \bibinfo{volume}{6}~(\bibinfo{number}{6})
  (\bibinfo{year}{1995}) \bibinfo{pages}{1320--1329}.

\bibitem[{Zhang and Suganthan(2016{\natexlab{a}})}]{zhang2016comprehensive}
\bibinfo{author}{L.~Zhang}, \bibinfo{author}{P.~N. Suganthan},
  \bibinfo{title}{A comprehensive evaluation of random vector functional link
  networks}, \bibinfo{journal}{Information sciences} \bibinfo{volume}{367}
  (\bibinfo{year}{2016}{\natexlab{a}}) \bibinfo{pages}{1094--1105}.

\bibitem[{Vovk(2013)}]{vovk2013kernel}
\bibinfo{author}{V.~Vovk}, \bibinfo{title}{Kernel ridge regression}, in:
  \bibinfo{booktitle}{Empirical inference}, \bibinfo{publisher}{Springer},
  \bibinfo{pages}{105--116}, \bibinfo{year}{2013}.

\bibitem[{Saunders et~al.(1998)Saunders, Gammerman, and
  Vovk}]{saunders1998ridge}
\bibinfo{author}{C.~Saunders}, \bibinfo{author}{A.~Gammerman},
  \bibinfo{author}{V.~Vovk}, \bibinfo{title}{Ridge regression learning
  algorithm in dual variables.}, in: \bibinfo{booktitle}{ICML},
  \bibinfo{pages}{515--521}, \bibinfo{year}{1998}.

\bibitem[{Bartlett and Mendelson(2002)}]{bartlett2002rademacher}
\bibinfo{author}{P.~L. Bartlett}, \bibinfo{author}{S.~Mendelson},
  \bibinfo{title}{Rademacher and Gaussian complexities: Risk bounds and
  structural results}, \bibinfo{journal}{Journal of Machine Learning Research}
  \bibinfo{volume}{3}~(\bibinfo{number}{Nov}) (\bibinfo{year}{2002})
  \bibinfo{pages}{463--482}.

\bibitem[{Bishop(2006)}]{bishop2006pattern}
\bibinfo{author}{C.~M. Bishop}, \bibinfo{title}{Pattern recognition and machine
  learning}, \bibinfo{publisher}{springer}, \bibinfo{year}{2006}.

\bibitem[{Xu et~al.(2017)Xu, Li, and Yang}]{xu2017kernel}
\bibinfo{author}{K.-K. Xu}, \bibinfo{author}{H.-X. Li}, \bibinfo{author}{H.-D.
  Yang}, \bibinfo{title}{Kernel-Based Random Vector Functional-Link Network for
  Fast Learning of Spatiotemporal Dynamic Processes}, \bibinfo{journal}{IEEE
  Transactions on Systems, Man, and Cybernetics: Systems} .

\bibitem[{Shi et~al.(2018)Shi, Xue, Dai, Peng, Dong, Zhang, and
  Zhang}]{shi2018cascaded}
\bibinfo{author}{J.~Shi}, \bibinfo{author}{Z.~Xue}, \bibinfo{author}{Y.~Dai},
  \bibinfo{author}{B.~Peng}, \bibinfo{author}{Y.~Dong},
  \bibinfo{author}{Q.~Zhang}, \bibinfo{author}{Y.~Zhang},
  \bibinfo{title}{Cascaded Multi-Column RVFL+ Classifier for Single-Modal
  Neuroimaging-Based Diagnosis of Parkinson's Disease}, \bibinfo{journal}{IEEE
  Transactions on Biomedical Engineering} .

\bibitem[{Dai et~al.(2018)Dai, Dong, Zhang, and Zhang}]{dai2018transcranial}
\bibinfo{author}{Y.~Dai}, \bibinfo{author}{Y.~Dong},
  \bibinfo{author}{Q.~Zhang}, \bibinfo{author}{Y.~Zhang},
  \bibinfo{title}{Transcranial Sonography Based Diagnosis Of Parkinson’s
  Disease Via Cascaded Kernel RVFL+}, in: \bibinfo{booktitle}{2018 40th Annual
  International Conference of the IEEE Engineering in Medicine and Biology
  Society (EMBC)}, \bibinfo{organization}{IEEE}, \bibinfo{pages}{574--577},
  \bibinfo{year}{2018}.

\bibitem[{Liu and Fieguth(2012)}]{liu2012texture}
\bibinfo{author}{L.~Liu}, \bibinfo{author}{P.~Fieguth}, \bibinfo{title}{Texture
  classification from random features}, \bibinfo{journal}{IEEE Transactions on
  Pattern Analysis and Machine Intelligence}
  \bibinfo{volume}{34}~(\bibinfo{number}{3}) (\bibinfo{year}{2012})
  \bibinfo{pages}{574--586}.

\bibitem[{Breiman(2001)}]{breiman2001random}
\bibinfo{author}{L.~Breiman}, \bibinfo{title}{Random forests},
  \bibinfo{journal}{Machine learning}
  \bibinfo{volume}{45}~(\bibinfo{number}{1}) (\bibinfo{year}{2001})
  \bibinfo{pages}{5--32}.

\bibitem[{Breiman(1996)}]{breiman1996bagging}
\bibinfo{author}{L.~Breiman}, \bibinfo{title}{Bagging predictors},
  \bibinfo{journal}{Machine learning}
  \bibinfo{volume}{24}~(\bibinfo{number}{2}) (\bibinfo{year}{1996})
  \bibinfo{pages}{123--140}.

\bibitem[{Wang and Li(2017)}]{wang2017stochastic}
\bibinfo{author}{D.~Wang}, \bibinfo{author}{M.~Li}, \bibinfo{title}{Stochastic
  configuration networks: Fundamentals and algorithms}, \bibinfo{journal}{IEEE
  transactions on cybernetics} \bibinfo{volume}{47}~(\bibinfo{number}{10})
  (\bibinfo{year}{2017}) \bibinfo{pages}{3466--3479}.

\bibitem[{Zhu et~al.(2019)Zhu, Feng, Wang, Jia, and He}]{zhu2019further}
\bibinfo{author}{X.~Zhu}, \bibinfo{author}{X.~Feng}, \bibinfo{author}{W.~Wang},
  \bibinfo{author}{X.~Jia}, \bibinfo{author}{R.~He}, \bibinfo{title}{A Further
  Study on the Inequality Constraints in Stochastic Configuration Networks},
  \bibinfo{journal}{Information Sciences} .

\bibitem[{Cao et~al.(2018)Cao, Wang, Ming, and Gao}]{cao2018review}
\bibinfo{author}{W.~Cao}, \bibinfo{author}{X.~Wang}, \bibinfo{author}{Z.~Ming},
  \bibinfo{author}{J.~Gao}, \bibinfo{title}{A review on neural networks with
  random weights}, \bibinfo{journal}{Neurocomputing} \bibinfo{volume}{275}
  (\bibinfo{year}{2018}) \bibinfo{pages}{278--287}.

\bibitem[{Zhang and Suganthan(2016{\natexlab{b}})}]{zhang2016survey}
\bibinfo{author}{L.~Zhang}, \bibinfo{author}{P.~N. Suganthan},
  \bibinfo{title}{A survey of randomized algorithms for training neural
  networks}, \bibinfo{journal}{Information Sciences} \bibinfo{volume}{364}
  (\bibinfo{year}{2016}{\natexlab{b}}) \bibinfo{pages}{146--155}.

\bibitem[{Chen and Wan(1999)}]{chen1999rapid}
\bibinfo{author}{C.~P. Chen}, \bibinfo{author}{J.~Z. Wan}, \bibinfo{title}{A
  rapid learning and dynamic stepwise updating algorithm for flat neural
  networks and the application to time-series prediction},
  \bibinfo{journal}{IEEE Transactions on Systems, Man, and Cybernetics, Part B
  (Cybernetics)} \bibinfo{volume}{29}~(\bibinfo{number}{1})
  (\bibinfo{year}{1999}) \bibinfo{pages}{62--72}.

\bibitem[{Chen(1996)}]{chen1996rapid}
\bibinfo{author}{C.~P. Chen}, \bibinfo{title}{A rapid supervised learning
  neural network for function interpolation and approximation},
  \bibinfo{journal}{IEEE Transactions on Neural Networks}
  \bibinfo{volume}{7}~(\bibinfo{number}{5}) (\bibinfo{year}{1996})
  \bibinfo{pages}{1220--1230}.

\bibitem[{Patra et~al.(1999)Patra, Pal, Chatterji, and
  Panda}]{patra1999identification}
\bibinfo{author}{J.~C. Patra}, \bibinfo{author}{R.~N. Pal},
  \bibinfo{author}{B.~Chatterji}, \bibinfo{author}{G.~Panda},
  \bibinfo{title}{Identification of nonlinear dynamic systems using functional
  link artificial neural networks}, \bibinfo{journal}{IEEE Transactions on
  Systems, Man, and Cybernetics, Part B (Cybernetics)}
  \bibinfo{volume}{29}~(\bibinfo{number}{2}) (\bibinfo{year}{1999})
  \bibinfo{pages}{254--262}.

\bibitem[{Cui et~al.(2018)Cui, Zhang, Li, Guo, Meng, Wang, and
  Xie}]{cui2018received}
\bibinfo{author}{W.~Cui}, \bibinfo{author}{L.~Zhang}, \bibinfo{author}{B.~Li},
  \bibinfo{author}{J.~Guo}, \bibinfo{author}{W.~Meng},
  \bibinfo{author}{H.~Wang}, \bibinfo{author}{L.~Xie}, \bibinfo{title}{Received
  signal strength based indoor positioning using a random vector functional
  link network}, \bibinfo{journal}{IEEE Transactions on Industrial Informatics}
  \bibinfo{volume}{14}~(\bibinfo{number}{5}) (\bibinfo{year}{2018})
  \bibinfo{pages}{1846--1855}.

\bibitem[{Zhang et~al.(2019)Zhang, Wu, Cai, Du, and
  Philip}]{zhang2019unsupervised}
\bibinfo{author}{Y.~Zhang}, \bibinfo{author}{J.~Wu}, \bibinfo{author}{Z.~Cai},
  \bibinfo{author}{B.~Du}, \bibinfo{author}{S.~Y. Philip}, \bibinfo{title}{An
  unsupervised parameter learning model for RVFL neural network},
  \bibinfo{journal}{Neural Networks} \bibinfo{volume}{112}
  (\bibinfo{year}{2019}) \bibinfo{pages}{85--97}.

\bibitem[{Scardapane et~al.(2018)Scardapane, Wang, and
  Uncini}]{scardapane2018bayesian}
\bibinfo{author}{S.~Scardapane}, \bibinfo{author}{D.~Wang},
  \bibinfo{author}{A.~Uncini}, \bibinfo{title}{Bayesian random vector
  functional-link networks for robust data modeling}, \bibinfo{journal}{IEEE
  transactions on cybernetics} \bibinfo{volume}{48}~(\bibinfo{number}{7})
  (\bibinfo{year}{2018}) \bibinfo{pages}{2049--2059}.

\bibitem[{Pratama et~al.(2018)Pratama, Angelov, Lughofer, and
  Er}]{pratama2018parsimonious}
\bibinfo{author}{M.~Pratama}, \bibinfo{author}{P.~P. Angelov},
  \bibinfo{author}{E.~Lughofer}, \bibinfo{author}{M.~J. Er},
  \bibinfo{title}{Parsimonious random vector functional link network for data
  streams}, \bibinfo{journal}{Information Sciences} \bibinfo{volume}{430}
  (\bibinfo{year}{2018}) \bibinfo{pages}{519--537}.

\bibitem[{Chen and Liu(2018)}]{chen2018broad}
\bibinfo{author}{C.~P. Chen}, \bibinfo{author}{Z.~Liu}, \bibinfo{title}{Broad
  learning system: An effective and efficient incremental learning system
  without the need for deep architecture}, \bibinfo{journal}{IEEE transactions
  on neural networks and learning systems}
  \bibinfo{volume}{29}~(\bibinfo{number}{1}) (\bibinfo{year}{2018})
  \bibinfo{pages}{10--24}.

\bibitem[{Chen and Zhang(2014)}]{chen2014data}
\bibinfo{author}{C.~P. Chen}, \bibinfo{author}{C.-Y. Zhang},
  \bibinfo{title}{Data-intensive applications, challenges, techniques and
  technologies: A survey on Big Data}, \bibinfo{journal}{Information Sciences}
  \bibinfo{volume}{275} (\bibinfo{year}{2014}) \bibinfo{pages}{314--347}.

\bibitem[{Giryes et~al.(2016)Giryes, Sapiro, and Bronstein}]{giryes2016deep}
\bibinfo{author}{R.~Giryes}, \bibinfo{author}{G.~Sapiro},
  \bibinfo{author}{A.~M. Bronstein}, \bibinfo{title}{Deep neural networks with
  random Gaussian weights: a universal classification strategy?},
  \bibinfo{journal}{IEEE Trans. Signal Processing}
  \bibinfo{volume}{64}~(\bibinfo{number}{13}) (\bibinfo{year}{2016})
  \bibinfo{pages}{3444--3457}.

\bibitem[{Wolf and Shashua(2003)}]{wolf2003learning}
\bibinfo{author}{L.~Wolf}, \bibinfo{author}{A.~Shashua},
  \bibinfo{title}{Learning over sets using kernel principal angles},
  \bibinfo{journal}{Journal of Machine Learning Research}
  \bibinfo{volume}{4}~(\bibinfo{number}{Oct}) (\bibinfo{year}{2003})
  \bibinfo{pages}{913--931}.

\bibitem[{Glorot and Bengio(2010)}]{glorot2010understanding}
\bibinfo{author}{X.~Glorot}, \bibinfo{author}{Y.~Bengio},
  \bibinfo{title}{Understanding the difficulty of training deep feedforward
  neural networks}, in: \bibinfo{booktitle}{Proceedings of the thirteenth
  international conference on artificial intelligence and statistics},
  \bibinfo{pages}{249--256}, \bibinfo{year}{2010}.

\bibitem[{He et~al.(2015)He, Zhang, Ren, and Sun}]{he2015delving}
\bibinfo{author}{K.~He}, \bibinfo{author}{X.~Zhang}, \bibinfo{author}{S.~Ren},
  \bibinfo{author}{J.~Sun}, \bibinfo{title}{Delving deep into rectifiers:
  Surpassing human-level performance on imagenet classification}, in:
  \bibinfo{booktitle}{Proceedings of the IEEE international conference on
  computer vision}, \bibinfo{pages}{1026--1034}, \bibinfo{year}{2015}.

\bibitem[{Bartlett(1998)}]{bartlett1998sample}
\bibinfo{author}{P.~L. Bartlett}, \bibinfo{title}{The sample complexity of
  pattern classification with neural networks: the size of the weights is more
  important than the size of the network}, \bibinfo{journal}{IEEE transactions
  on Information Theory} \bibinfo{volume}{44}~(\bibinfo{number}{2})
  (\bibinfo{year}{1998}) \bibinfo{pages}{525--536}.

\bibitem[{Cortes and Vapnik(1995)}]{cortes1995support}
\bibinfo{author}{C.~Cortes}, \bibinfo{author}{V.~Vapnik},
  \bibinfo{title}{Support vector machine}, \bibinfo{journal}{Machine learning}
  \bibinfo{volume}{20}~(\bibinfo{number}{3}) (\bibinfo{year}{1995})
  \bibinfo{pages}{273--297}.

\bibitem[{Suykens and Vandewalle(1999)}]{suykens1999least}
\bibinfo{author}{J.~A. Suykens}, \bibinfo{author}{J.~Vandewalle},
  \bibinfo{title}{Least squares support vector machine classifiers},
  \bibinfo{journal}{Neural processing letters}
  \bibinfo{volume}{9}~(\bibinfo{number}{3}) (\bibinfo{year}{1999})
  \bibinfo{pages}{293--300}.

\bibitem[{Kakade et~al.(2009)Kakade, Sridharan, and
  Tewari}]{kakade2009complexity}
\bibinfo{author}{S.~M. Kakade}, \bibinfo{author}{K.~Sridharan},
  \bibinfo{author}{A.~Tewari}, \bibinfo{title}{On the complexity of linear
  prediction: Risk bounds, margin bounds, and regularization}, in:
  \bibinfo{booktitle}{Advances in neural information processing systems},
  \bibinfo{pages}{793--800}, \bibinfo{year}{2009}.

\bibitem[{LeCun et~al.(1998)LeCun, Bottou, Bengio, and
  Haffner}]{lecun1998gradient}
\bibinfo{author}{Y.~LeCun}, \bibinfo{author}{L.~Bottou},
  \bibinfo{author}{Y.~Bengio}, \bibinfo{author}{P.~Haffner},
  \bibinfo{title}{Gradient-based learning applied to document recognition},
  \bibinfo{journal}{Proceedings of the IEEE}
  \bibinfo{volume}{86}~(\bibinfo{number}{11}) (\bibinfo{year}{1998})
  \bibinfo{pages}{2278--2324}.

\bibitem[{Bergstra and Bengio(2012)}]{bergstra2012random}
\bibinfo{author}{J.~Bergstra}, \bibinfo{author}{Y.~Bengio},
  \bibinfo{title}{Random search for hyper-parameter optimization},
  \bibinfo{journal}{Journal of Machine Learning Research}
  \bibinfo{volume}{13}~(\bibinfo{number}{Feb}) (\bibinfo{year}{2012})
  \bibinfo{pages}{281--305}.

\bibitem[{Chang and Lin(2011)}]{CC01a}
\bibinfo{author}{C.-C. Chang}, \bibinfo{author}{C.-J. Lin},
  \bibinfo{title}{{LIBSVM}: A library for support vector machines},
  \bibinfo{journal}{ACM Transactions on Intelligent Systems and Technology}
  \bibinfo{volume}{2} (\bibinfo{year}{2011}) \bibinfo{pages}{27:1--27:27},
  \bibinfo{note}{software available at
  \url{http://www.csie.ntu.edu.tw/~cjlin/libsvm}}.

\bibitem[{Liang and Cherkassky(2008)}]{liang2008connection}
\bibinfo{author}{L.~Liang}, \bibinfo{author}{V.~Cherkassky},
  \bibinfo{title}{Connection between SVM+ and multi-task learning}, in:
  \bibinfo{booktitle}{Neural Networks, 2008. IJCNN 2008.(IEEE World Congress on
  Computational Intelligence). IEEE International Joint Conference on},
  \bibinfo{organization}{IEEE}, \bibinfo{pages}{2048--2054},
  \bibinfo{year}{2008}.

\bibitem[{Lichman(2013)}]{Lichman:2013}
\bibinfo{author}{M.~Lichman}, \bibinfo{title}{{UCI} Machine Learning
  Repository}, \urlprefix\url{http://archive.ics.uci.edu/ml},
  \bibinfo{year}{2013}.

\bibitem[{Spyromitros-Xioufis et~al.(2016)Spyromitros-Xioufis, Tsoumakas,
  Groves, and Vlahavas}]{spyromitros2016multi}
\bibinfo{author}{E.~Spyromitros-Xioufis}, \bibinfo{author}{G.~Tsoumakas},
  \bibinfo{author}{W.~Groves}, \bibinfo{author}{I.~Vlahavas},
  \bibinfo{title}{Multi-target regression via input space expansion: treating
  targets as inputs}, \bibinfo{journal}{Machine Learning}
  \bibinfo{volume}{104}~(\bibinfo{number}{1}) (\bibinfo{year}{2016})
  \bibinfo{pages}{55--98}.

\end{thebibliography}





\end{document}